\documentclass{article}

\PassOptionsToPackage{numbers, compress}{natbib}

\usepackage[preprint]{neurips_2024}

\usepackage{booktabs}
\usepackage{adjustbox}
\usepackage{rotating}

\usepackage[utf8]{inputenc} 
\usepackage[T1]{fontenc}    
\usepackage{hyperref}       
\usepackage{url}            
\usepackage{booktabs}       
\usepackage{amsfonts}       
\usepackage{nicefrac}       
\usepackage{microtype}      
\usepackage[table]{xcolor}  

\usepackage{longtable}
\usepackage{multicol}
\usepackage{multirow}
\usepackage{adjustbox}
\usepackage{amsmath}
\usepackage{comment}
\usepackage{enumitem} 
\usepackage{tabularx}
\usepackage{graphicx}

\usepackage{todonotes}
\usepackage{setspace} 
\let\oldtodo\todo
\renewcommand{\todo}[1]{\oldtodo{\begin{spacing}{0.5}\tiny #1\end{spacing}}} 
\definecolor{my_green}{RGB}{51,102,0}
\definecolor{my_yellow}{RGB}{255,255,0}
\definecolor{my_red}{RGB}{204, 0, 0}
\usepackage{pifont} 
\newcommand{\cmark}{\textcolor{my_green}{\ding{51}}} 
\newcommand{\xmark}{\textcolor{my_red}{\ding{55}}} 

\newcommand{\ppname}{Co-STEER}
\newcommand{\sname}{AD\textsuperscript{2}}

\title{Collaborative Evolving Strategy for Automatic Data-Centric Development}

\author {
    Xu Yang\textsuperscript{\rm 2}\footnotemark[1] , Haotian Chen\textsuperscript{\rm 1}\thanks{Equally Contributed} , Wenjun Feng\textsuperscript{\rm 1}\footnotemark[1] , Haoxue Wang\textsuperscript{\rm 1}, Zeqi Ye\textsuperscript{\rm 1},  Xinjie Shen\textsuperscript{\rm 1}  \\
    \textbf{Xiao Yang\textsuperscript{\rm 2}\thanks{Corresponding Author} , Shizhao Sun\textsuperscript{\rm 2},  Weiqing Liu\textsuperscript{\rm 2}, Jiang Bian\textsuperscript{\rm 2}} \\
    \textsuperscript{\rm 2}Microsoft Research Asia \\
    \texttt{xuyang1@microsoft.com}, \texttt{ht1ian.chen@gmail.com}, \\
    \texttt{fwj20020813@outlook.com}, \texttt{\{whx924, liamyzq, frinkleko\}@gmail.com} \\
    \texttt{\{xiao.yang, shizhao.sun, weiqing.liu, jiang.bian\}@microsoft.com}
}

\begin{document}

\maketitle

\begin{abstract}
Artificial Intelligence (AI) significantly influences many fields, largely thanks to the vast amounts of high-quality data for machine learning models. The emphasis is now on a data-centric AI strategy, prioritizing data development over model design progress. Automating this process is crucial.
In this paper, we serve as the first work to introduce the automatic data-centric development (\sname{}) task and outline its core challenges, which require domain-experts-like task scheduling and implementation capability, largely unexplored by previous work.
  By leveraging the strong complex problem-solving capabilities of large language models (LLMs), we propose an LLM-based autonomous agent, equipped with a strategy named Collaborative Knowledge-STudying-Enhanced Evolution by Retrieval (Co-STEER), to simultaneously address all the challenges. Specifically, our proposed Co-STEER agent enriches its domain knowledge through our proposed evolving strategy and develops both its scheduling and implementation skills by accumulating and retrieving domain-specific practical experience.
With an improved schedule, the capability for implementation accelerates. Simultaneously, as implementation feedback becomes more thorough, the scheduling accuracy increases. These two capabilities evolve together through practical feedback, enabling a collaborative evolution process.
 Extensive experimental results demonstrate that our Co-STEER agent breaks new ground in \sname{} research, possesses strong evolvable schedule and implementation ability, and demonstrates the significant effectiveness of its components. Our Co-STEER paves the way for \sname{} advancements.
\end{abstract}

\section{Introduction}

\renewcommand{\thefootnote}{}
\footnotetext{This is an open-source project starting in Oct. 2023.  \textsuperscript{\rm 1} Work done during an internship at Microsoft.}
\renewcommand{\thefootnote}{\arabic{footnote}}

Scientific advances have proceeded via a combination of different paradigms~\cite{hey2009the,tao2024report}, where data-driven discovery is an emerging paradigm formalized in recent years and more significantly accelerates scientific advances~\cite{hey2009the}. All the paradigms experience impediments to progress, including the expensive time cost of verifying a scientific hypothesis, the extremely vast and complex range of candidate theories, and the requirements of enormous amounts of computational resources. Data-driven discovery particularly suffers from these impediments. It is a long-standing aspiration to reduce the impediments and accelerate the rate of scientific progress~\cite{wang2023scientific}. With the recent advances in AI, especially in AI agents based on LLMs~\cite{devlin2018bert,wei2022chain,mialon2023augmented,openai2023gpt4,durante2024an}, the aspiration is more likely to become a reality~\cite{wang2023scientific}.

\begin{figure}[ht]
    \centering
    \includegraphics[width=1.0\textwidth]{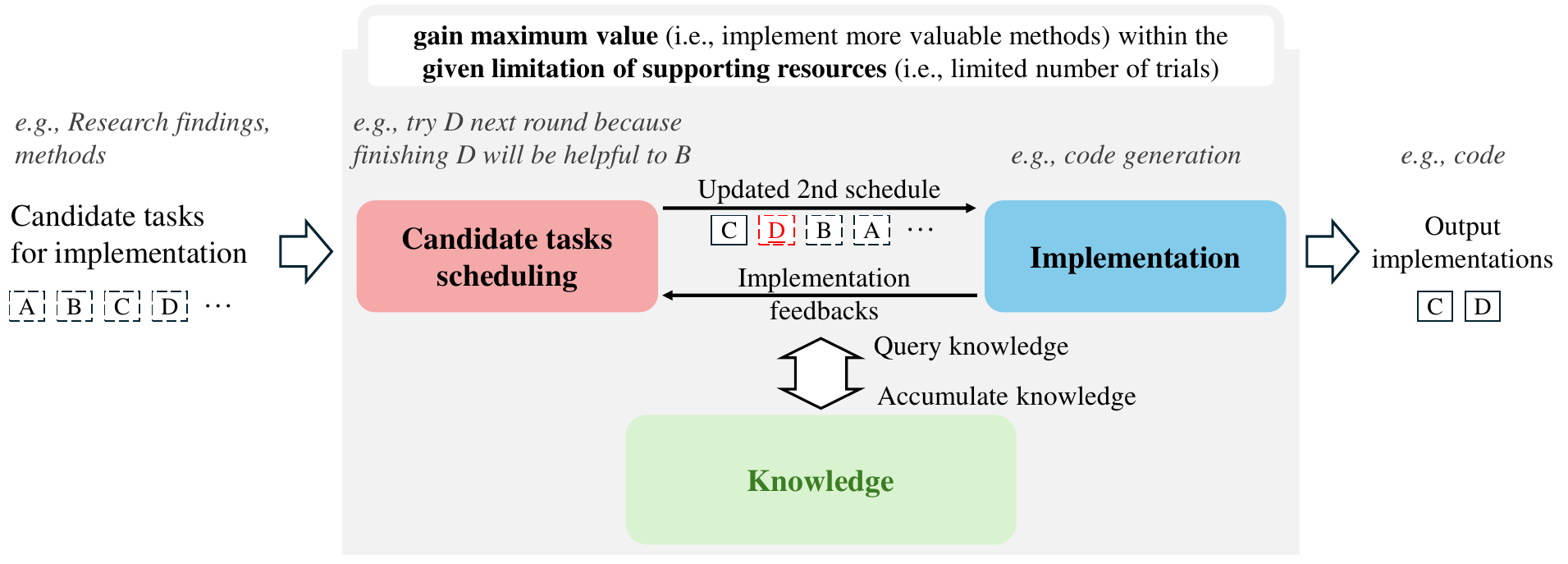}
    \caption{A brief illustration of \sname{}. An agent is expected to understand both the current method and candidate data sources for data selection and preprocessing in the engineering logic expression step.}
    \label{fig:intro}
\end{figure}

In this paper, we serve as the first effort to \textbf{A}utomate \textbf{D}ata-centric \textbf{D}evelopment (\sname{})~\cite{zha2023data} with the aid of LLM-based agents, propelling the data-driven discovery paradigm forward.
Specifically, we take the first step toward tackling a representative yet unexplored real-world scenario of automatic data-centric development , where agents are expected to substitute human researchers to first \textbf{schedule} to prioritize candidate methods (solutions) due to the enormous number of candidates and the limited computational resources, then prepare the appropriate engineering \textbf{implementation} of the prioritized methods, and finally execute their implementations to obtain accurate results. 
Each candidate method in \sname{}  is a data development task, such as feature extraction.  A typical example of this is the implementation of financial factors, a domain where numerous researchers publish their research finding about developing advanced financial datasets~\cite{kakushadze2016101}.
The common \sname{} scenario is shown in Figure~\ref{fig:intro}. The distinctive challenges of \sname{} and our corresponding contributions are as follows.


The first challenge is that \sname{} requires agents to be more efficient by scheduling to prioritize the vast array of candidate methods to gain maximum value (i.e., realizing more valuable methods) within the given limitation of supporting resources.
For example, candidate methods, including hundreds of thousands of chip designs, compounds for medicine or materials, and model architectures are waiting for validation from human experts~\cite{bloom2020are,mock2023ai,zeni2024mattergen,liu2024chipnemo}. 
The experimental results of certain candidates will provide crucial information that supports the implementation of others: CoT~\cite{wei2022chain} inspired other methods such as ReAct~\cite{yao2023react} and Tree-of-Thought~\cite{yao2023tree}.
Therefore, an effective schedule can streamline the implementation process, ensuring the successful delivery of more implementations.
Furthermore, a significant discovery can effectively guide the research agenda, enabling better prioritization of tasks that yield more significant results.
Previous work~\cite{qin2023tool,mialon2023augmented,schick2023toolformer,yuan2024craft} focuses on implementation while rarely considering scheduling them.
\textbf{We argue that an effective \sname{} agent can 1) schedule to accomplish more implementation tasks within limited resources 2) and evolve through practical feedback like human experts.}



The second challenge lies in the fact that \sname{} tasks are fundamentally research implementation tasks, which are significantly more complex than standard coding tasks.
1) These tasks demand that agents are equipped with a higher level of domain-specific knowledge, which in turn bolsters their implementation capabilities.
Coding agents based on LLM typically incorporate general coding datasets during training.
However, these datasets often lack the specific knowledge required for \sname{} tasks, resulting in subpar performance in these complex tasks.
Previous works boost the general implementation ability of LLM agents by adopting code-specific pretraining, self-correction, and planning strategies~\cite{yang2023failures,xu2023lemur,gou2024critic,wu2024oscopilot,lu2024self}.
However, they overlook the importance of enhancing the domain-specific knowledge of LLM agents, which is typically acquired through practical experience in specific scenarios.
For example, a programmer may neglect the unique characteristics when he starts in a new domain-specific scenario. With repeated practice, he gains expertise and is able to deliver effective solutions.
2) On the other hand, simply incorporating knowledge through demonstrations in the learning context may not suffice for complex \sname{} tasks. Much like real-world development, developers often solve a series of problems through multiple iterations before delivering a solution. Finding a similar demonstration that aligns with the current series of problems can be challenging.
This presents a challenge in effectively transferring previously acquired knowledge to current, complex tasks.
\textbf{We argue that a qualified \sname{} agent should 1) continuously learn domain knowledge from practice and 2) then effectively use it. }

The potential ideal agents, capable of tackling the two challenges, share a common philosophy: evolving through practice. 
Specifically, accumulating a knowledge base through practice strengthens both the scheduling and implementation abilities of agents.
Meanwhile, a proficient scheduling agent enhances its scheduling strategies by integrating valuable feedback from practical implementations. Likewise, an implementation agent accelerates its evolution by leveraging sensible schedules.
Thus, an effective \sname{} solution should progress through cooperative evolution.

To address these challenges, we propose an LLM-based automatic developing agent, equipped with a strategy named Collaborative Knowledge-STudying-Enhanced Evolution by Retrieval (Co-STEER). Co-STEER simulates the career progression of an engineer from junior to senior level: The abilities of agents evolve through their own knowledge.
Specifically, Co-STEER offers a co-evolving solution through a scheduling agent and an implementation agent. The scheduling agent evolves by considering the dependencies of candidate methods and learning from implementation feedback, thereby creating more effective schedules. The implementation agent, on the other hand, evolves through practical experience, developing a transferrable knowledge base for complex tasks such as \sname{}. As these agents evolve, they share feedback and status updates with each other, fostering collaborative evolution.
To highlight our innovative approach, we have compared our method, \ppname{}, with previous works in the field of natural-language-to-code. The comparison is presented in Table \ref{tab:method_comparison}.

Experiments have been conducted on a typical \sname{} benchmark dataset, financial factors implementation, demonstrating the effectiveness of our proposed \ppname{} agent.
The implementation correctness outperforms previous SOTA methods by 44.5\% (from 0.454 to 0.646). With the assistance of the scheduling agent, performance has been boosted by 24\% on average in different settings.

\begin{table*}[t!]
 \centering
 \vspace{-2mm}
 \small
 \renewcommand\tabcolsep{3.5pt} 
\renewcommand\arraystretch{0.80} 
 \resizebox{1.0\linewidth}{!}{
 \begin{tabular}{lc|ccccc} 
 \toprule
 \multirow{2}{*}{Methods} & 
 \multicolumn{1}{c}{Schedule} & 
 \multicolumn{5}{c}{Implementation} \\ 
 \cmidrule(lr){2-2} \cmidrule(l){3-7}
            & - & Demonstration & Planning or Reasoning Before Implementation & LLM-Based Self-Feedback & External Feedback & Growing Practical Knowledge \\
  \cmidrule(r){1-1} 
  \cmidrule(lr){2-2} \cmidrule(lr){3-3} \cmidrule(lr){4-4} \cmidrule(lr){5-5} \cmidrule(lr){6-6} \cmidrule(l){7-7}

 Few-shot~\cite{brown2020lanugage} & \xmark & \cmark & \xmark & \xmark & \xmark & \xmark \\
 CoT~\cite{ma2023chain} & \xmark & \xmark & \cmark & \xmark & \xmark & \xmark \\
 Reflexion~\cite{shinn2023reflexion} & \xmark & \xmark & \xmark & \cmark & \cmark & \xmark \\
 Self-Debugging~\cite{jiang2023self} & \xmark & \xmark & \xmark & \cmark & \cmark & \xmark \\
 Self-Planning~\cite{chen2024teaching} & \xmark & \xmark & \cmark & \xmark & \xmark & \xmark \\
 \textbf{\ppname{}} & \cellcolor{my_yellow} \cmark & \cmark & \cmark & \cmark & \cmark & \cellcolor{my_yellow}\cmark \\
\bottomrule
 \end{tabular}
}
\caption{This table contrasts various natural-language-to-code methodologies, highlighting \ppname{} as the pioneering approach that encompasses the entire \sname{} workflow, from scheduling to implementation. Compared with prior state-of-the-art solutions, \ppname{} introduces a growing practical knowledge base, significantly enhancing learnability in the domain.}
 \vspace{-2mm}
 \label{tab:method_comparison}
\end{table*}

\section{Related Work}
    


\subsection{Agent Workflows}
LLM-based agents refer to the software agents that leverage the capabilities of LLMs to perform a wide range of tasks.
Our method, which focuses on the data-centric development problem, belongs to the big family of LLM-based agents.
In the following, we introduce the mainstream techniques frequently used by LLM-based agents, and the differences between those techniques and our method.

\noindent\textbf{Planning.}
Planning is critical to achieving predefined goals. 
CoT~\cite{wei2022chain} proposes to decouple a task into several steps and make LLM follow these steps by prompting.
Subsequent methods further automate the process of decoupling a task~\cite{zhang2023automatic}, integrate the actions into reasoning~\cite{yao2023react} and introduce multiple reasoning paths~\cite{yao2023tree}.
These studies focus on decomposing a task into steps.
Different from them, we put effort into prioritizing candidate steps, which is a critical part of planning but receives little attention.

\noindent\textbf{Self-correction.}
Self-correction aims at improving an agent's performance by assessing the quality of LLM-generated results and giving feedback to LLM.
Feedback could be one-time~\cite{saunders2022selfcritiquing,chen2024teaching} or in an iterative manner~\cite{ridnik2024code,lu2024self,gou2024critic}, and it could come from LLMs~\cite{yang2023failures,an2024learning} or existing knowledge~\cite{peng2023semiparametric,peng2023check}.
Our method leverages the main idea of self-correction (i.e., using feedback), but is significantly different from existing work.
Specifically, we accumulate practical experience from real experiments to build a knowledge base and leverage the knowledge base to give feedback, while existing studies only consider off-the-shelf domain knowledge from textbooks or even do not have a knowledge base. 
Moreover, we design a special mechanism to make the knowledge transferable among different tasks.

\noindent\textbf{Tool learning.}
Tool learning refers to the process by which an LLM learns to use external tools or resources to enhance its problem-solving capabilities.
Recent efforts focus on teaching LLMs to be proficient with existing tools either by fine-tuning~\cite{qin2024toolllm,schick2023toolformer,li2024toolaugmented} or tuning-free methods~\cite{yuan2024craft,wang2023voyager,zhang2024training,gao2024clova}.
Differently, we consider a more advanced way of tool learning.
First, instead of simply calling APIs from existing tools, we utilize and combine given tools to build a new tool.
Second, we leverage the knowledge base, especially the accumulated practical experience in it, to help tool learning.
\subsection{Agents in Related Scenarios}
Recently, various LLM-based agents have been developed to solve problems in different scenarios.
In the following, we discuss the relationships of those agents to our proposed method.

\noindent\textbf{Scientific research.}
Scientific research is a complex process consisting of two key components: the formulation of a new research idea and the development of the proposed idea.
A recent work~\cite{boiko2023emergent} shows the potential of LLM in scientific research by case studies with human evaluation.
A subsequent work~\cite{baek2024researchagent} presents a more systematic way to leverage LLM for idea formulation.
In this work, we delve into another crucial component of scientific research which has not been well studied yet, i.e., the development problem, especially the data-centric one.

\noindent\textbf{Machine learning.}
Recently, there has been a growing interest in designing an agent to automate machine learning.
CAAFE~\cite{hollmann2024large} focuses on the feature engineering problem, MLCopilot~\cite{zhang2023mlcopilot} investigates hyper-parameter tuning, and MLAgentBench~\cite{huang2023benchmarking} introduces 13 machine learning tasks and compares agents based on different LLMs.
In this work, we focus on a different task compared to machine learning, i.e., data-centric development, which faces distinct challenges.
Specifically, the key challenges of our task lies in prioritizing candidate methods and preparing engineering implementation, while the agents for machine learning tackle the challenge of automating specific components of the machine learning pipeline, e.g., feature engineering, hyper-parameter tuning and architecture search.

\noindent\textbf{Software development.}
To assist software development, LLM-based coding assistants, e.g., GitHub
Copilot~\cite{githubcopilot}, have advanced into
integrated development environments (IDEs).
Besides, AutoDev~\cite{tufano2024autodev} develops more powerful agents which can perform diverse operations on a codebase, including file editing, build processes, execution, testing, and git operations.
Unlike traditional software development, our focus is on data-driven development in the context of scientific research.
Specifically, the challenge of our task includes both prioritization and implementation while software development only considers implementation.
Moreover, even for implementation, an evolving knowledge base is indispensable in our method, while its role in the software development agent is not clearly discussed.

\section{Co-STEER Agent}

\subsection{Problem Formulation}
To get a clear understanding of the problem we are focusing on, as shown in Figure \ref{fig:intro}, we'll start with a formal definition in this section.
Given raw textual information comprising the descriptions of $N$ candidate tasks for implementation (e.g., some methods to be implemented \& verified), the goal of an \sname{} agent is to deliver as many completed implemented tasks as possible.
Due to \sname{} tasks being fundamentally research implementation tasks, which are usually novel and usually can't be achieved by calling existing tools, the task has to be implemented by creating a solution through coding. Thus, the outcome of implementation would be code.
The evaluation will be based on the execution results of the code, comparing results implemented by agents with the ground truth. We will have a quality score for the solution.
The more completed tasks, the higher the quality of the completed tasks, and the more value the \sname{} agent can achieve.
But the supporting resources are limited, like the development in the real world. In our scenario, the agent has only a limited number of trials.
The goal of the system is to gain maximum value within the given limitations of supporting resources.

Each agent can have multiple rounds of trials to implement tasks. The knowledge gained from practice will be very important to minimize the cost of implementing a task. For example, after gaining knowledge, the agent may take fewer trials (costs) to complete the same task. So the order (that may be adjusted dynamically based on the feedback from practice) to complete tasks to gain knowledge and the strategy to gain and leverage practical knowledge would be crucial for this scenario.

\subsection{Overall Design}
\begin{figure}[h]
    \centering
    \includegraphics[width=1.0\textwidth]{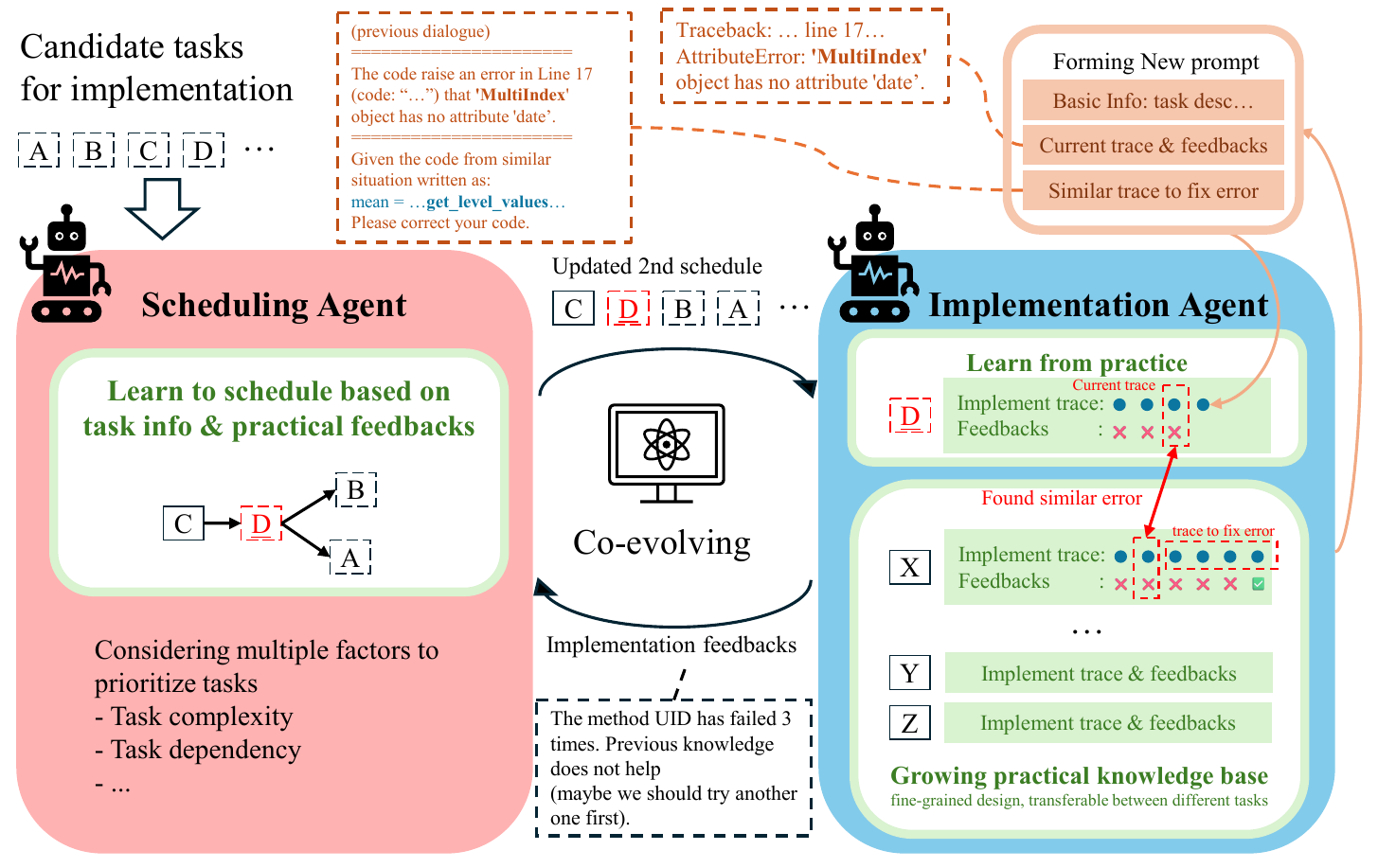}
    \caption{
    The detailed design of \ppname{} involves two agents: The scheduling agent inputs candidate tasks for implementation and tries to iteratively make schedules based on various factors; the implementation agent learns knowledge from practice and builds a fine-grained practical knowledge base that can transfer between different tasks; the agents evolve through mutual support.
  }
    \label{fig:method}
\end{figure}

Our proposed innovative approach, \ppname{}, presents a pioneering solution to address the problem outlined above. Illustrated in Figure \ref{fig:method}, the design integrates two core components: a scheduling agent and an implementation agent, each aiming to handle the challenge of scheduling and implementation, respectively.

The novelty of \ppname{} lies in its evolution through practice, which is crucial for achieving high performance in \sname{} tasks. It features a scheduling agent that refines its task schedules iteratively, drawing on feedback from practical implementations. Meanwhile, the implementation agent enhances its capabilities through ongoing practice, creating a knowledge base that can be transferred across various tasks.
The two agents, rather than operating in isolation, evolve collaboratively. The scheduling agent refines its task schedules by incorporating feedback from the implementation phase, leading to a more efficient schedule. Concurrently, the implementation agent evolves more smoothly under the optimized schedules provided by the scheduling agent, fostering a mutually beneficial growth dynamic.

\subsection{Scheduling Agent}
We'll introduce the design of the scheduling agent shown on the left in Figure \ref{fig:method}.
The scheduling agent plays a crucial role in determining the sequence of task attempts, thereby influencing the system's evolution.
On one hand, experimenting with diverse ideas not only garners valuable practical knowledge but also aids in the efficient completion of subsequent tasks, reducing overall costs. On the other hand, acquiring practical feedback enables the scheduling agent to deepen its comprehension of the task's essence, facilitating the formulation of a more effective schedule.

To effectively guide the scheduling agent towards its objectives, we have designed a workflow for the scheduling agent shown in Figure \ref{fig:method}. This workflow prompts the LLM to thoroughly evaluate a variety of factors when scheduling tasks. These factors include, but are not limited to, the complexity of the tasks and their interdependencies. We also encourage the LLM to take into account additional considerations that may impact the scheduling process.

Specifically, we introduce a \textit{guided Chain of Thought (CoT)} mechanism to facilitate the decision-making process.  This approach encourages the LLM to articulate its reasoning process before arriving at a final decision. Specifically, we prompt the LLM to map out the inherent Directed Acyclic Graph (DAG) among tasks, illustrating how the completion of one task may aid in accomplishing another. For instance, if acquiring the knowledge to complete task A could assist in completing task B, and task A is simpler than task B, we then establish a directed edge from A to B. We also request that the LLM provides a clear rationale for each directed edge it identifies.
This structured reasoning process, represented through the diagramming language Mermaid\footnote{\url{https://github.com/mermaid-js/mermaid}}, provides more reasoning guidance than the original CoT ~\cite{wei2022chain}.

Furthermore, the schedule is not static. The scheduling agent would collect feedback from the implementation of the tasks. This allows the scheduling agent to refine its understanding of each task's characteristics for more effective planning. For instance, repeated failures on the same task signal its difficulty, prompting the scheduler to prioritize simpler tasks to accumulate more knowledge efficiently. Please refer to Appendix \ref{sec:prompt} for detailed prompt design.

\subsection{Implementation Agent}
The implementation agent is on the right in Figure \ref{fig:method}.
A bad implementation agent needs more trials before getting a correct solution, which incurs greater cost.
Previous works boost the general implementation ability with a series of tricks, which are listed in Table \ref{tab:method_comparison}.
Our agent incorporates these widespread capabilities as well. However, the \sname{} tasks it tackles are fundamentally research-oriented implementation tasks, far surpassing the complexity of standard coding tasks. To effectively address these tasks, domain-specific knowledge is indispensable for devising viable solutions. Consequently, \ppname{} has been equipped with a growing practical knowledge base. This knowledge base is designed to continuously gather insights from practical experiences and transfer them to facilitate other tasks.

In the following section, we delve into the innovative aspects of this design.
The design of \ppname{}'s growing practical knowledge mainly focuses on the knowledge base and feedback.

\subsubsection{Knowledge Base Design}

The design of \ppname{}'s knowledge base aims to build a knowledge vault that can \textit{grow} and \textit{transfer}.

\noindent \textbf{Growing Practical Knowledge.} 
Practical knowledge represents the knowledge collected from practice by an agent in a specific domain. It aims to enhance the agent's capabilities in domain-specific tasks.
The design of the growing practical knowledge base is illustrated in the green box in Figure \ref{fig:method}. This knowledge base archives all successfully completed tasks. Each entry in the database not only records the final outcome but also documents the iterative process of trials and feedback encountered along the way. This detailed trace offers a wealth of information, highlighting various potential errors and the corresponding solutions to address them.
This approach records a comprehensive practical problem-solving path, far beyond just knowing the successful solution.

\noindent \textbf{Transferable Knowledge Usage.}
In addressing new tasks, incorporating examples of similar tasks into the prompt for in-context learning is a widely used strategy to apply past knowledge. However, as the complexity of tasks, such as those in \sname{}, increases, finding relevant examples becomes increasingly difficult. This limitation hinders the effectiveness of this common approach, making it challenging to transfer previously acquired knowledge to new, complex tasks.
\ppname{} introduces an innovative solution for complex tasks like \sname{}. As depicted on the right in Figure \ref{fig:method}, \ppname{} adopts a unique approach when encountering new tasks and errors (feedback). Instead of relying on task similarity, \ppname{} searches solutions based on feedback similarity within the practical knowledge base. The querying result is a list of steps to fix the current error. This detailed knowledge design and querying mechanism significantly increases the query hit rate and the success rate in resolving current errors, enabling the transfer of prior knowledge to address new, complex tasks.
For example, an agent based on GPT-4 erroneously uses a specific line of code expressed by ``index.date'' despite the feedback ``Attribution Error: MultiIndex has no attribution `date'''. By retrieving similar practical experience in other scenarios, the agent obtains a successful case where it uses ``df.index.get\_level\_values(`datetime').date'' to process the data. Such a retrieval result makes the agent successfully go through the current dilemma.
Note that we adopt an embedding model for similarity computation when executing queries.

\subsubsection{Feedback Design}
Feedback is essential for guiding agents towards the correct solutions. It plays a pivotal role in enabling agents to learn practical knowledge effectively. Therefore, it's imperative to design feedback that not only points agents in the right direction but also boosts their speed of evolution by learning from more informative practical knowledge.
Our evaluators are divided into two categories: the first operates autonomously, leveraging internal feedback mechanisms.
The second category requires ground truths crafted by human experts, providing more insightful feedback by comparing the current trial and the ground truth.

\noindent \textbf{Unsupervised Feedbacks.}
Unsupervised feedback is divided into two types: LLM-based and tool-based. LLM-based feedback, generated by LLM, evaluates the current solution and task, drawing insights similar to those in Reflexion~\cite{shinn2023reflexion}. On the other hand, tool-based feedback utilizes a Python compiler to run the proposed solutions, providing practical feedback on their execution.

\noindent \textbf{Supervised Feedbacks.}
In the presence of expert-crafted solutions, agents can enhance their learning by comparing their solutions and execution outcomes to these ground truths, thereby deriving deeper insights. The comparison process involves two key steps:
Firstly, for contrasting different solutions, LLMs are employed to succinctly articulate the distinctions between solutions generated by the agent and those crafted by experts. This step aids in identifying areas for improvement and refinement in the agent's solution.
Secondly, when evaluating execution results, feedback is grounded in quantitative measures such as the correlation coefficient and value accuracy, alongside other metrics.
This quantitative feedback provides insights from various perspectives, enriching the information available for improvement.

In the evolving process of the \ppname{}, it begins with an initial phase that leverages a small-sized knowledge base filled with expert-crafted solutions, ensuring the generation of high-quality feedback. This foundational knowledge base serves as a springboard for a warm start. As it progresses to address new tasks, the agent iteratively refines its approach through unsupervised feedback, continuously evolving and enhancing its implementation capabilities.

\section{Experiments}
In this section, we conduct experiments to answer the following research questions.


\begin{itemize}[itemsep=-0.5ex, parsep=0.5ex, leftmargin=0.12in]  
  \item \textbf{RQ1}: Does the implementation agent of \ppname{} outperform the previous (SOTA) natural-language-to-code baselines in \sname{} tasks?
  \item \textbf{RQ2}: Does the scheduling agent of \ppname{} make a reasonable schedule to boost the performance further?
\end{itemize}

\subsection{Datasets}
To verify the effectiveness of our method on \sname{} tasks, we conduct experiments in a typical \sname{} scenario: financial factor implementation. This domain is where numerous researchers publish their findings on developing advanced financial datasets~\cite{kakushadze2016101}. Daily financial research and development work involves scheduling and implementing factors across a vast array of candidate methods. To represent a real-world \sname{} scenario, we conduct experiments on RD2Bench~\cite{chen2024rdbench}, a benchmark consisting of 27 human-annotated implementable factors and 13 mistakenly extracted unimplementable factors. All factors are divided into three categories: fundamental, high-frequency, and price-volume factors. The difficulty of factors is categorized into three levels: easy, medium, and hard. The implementation of each category of factors requires different sources of data, which are provided in the benchmark. 

\subsection{Experimental Settings}


\paragraph{Baselines}
We include the following baselines in our experiments: Few-shot~\cite{brown2020lanugage}, CoT~\cite{wei2022chain}, Reflexion~\cite{shinn2023reflexion}, Self-Debugging~\cite{chen2024teaching}, and Self-Planning~\cite{jiang2023self}. Few-shot learning is a context learning method that enhances the model’s response formulation by providing several task-relevant examples and their answers. The Chain-of-Thought (CoT) model promotes logical progression in reasoning by necessitating step-by-step thinking. The Reflexion model, capable of introspection, identifies and corrects its own mistakes, thereby improving over time. The Self-Debugging model, through code analysis and feedback interpretation, can autonomously rectify errors. Lastly, the Self-Planning model demonstrates high autonomy by formulating its own action plan and making decisions or executing actions based on this plan.

\paragraph{Evaluation Metrics}

We include four evaluation metrics to evaluate the performance of the implementation agent: average execution, average format, average correlation, and maximum correlation.The average execution metric measures the average execution rate of the generated code; any error encountered during the execution will be counted as 0. The average format metric measures the average format correctness of the generated code, for example whether the column name is equalized as expected. The average correlation indicates the average correlation between the output series generated by the code generated by the model and the ground truth. For example, for the same input features, we evaluate the correlations of factors both produced by LLM's implementations. and the ground truth implementations.
The maximum correlation indicates the maximum correlation between the output series generated by the code generated by the model and the ground truth.

\subsection{Results of Method Implementation (RQ1)}

\begin{table}[!ht]
\centering
\adjustbox{width=0.6\textwidth}{%
\centering
\begin{tabular}{lcccc}
\toprule
 Methods &  avg. exec. & avg. format & avg. corr. & max. corr. \\
\midrule

\multirow[t]{4}{*}{Few-shot~\cite{brown2020lanugage}} &   0.733 & 0.433 & 0.454 & 0.562 \\

\multirow[t]{4}{*}{CoT~\cite{wei2022chain}} 
 & 0.833 & 0.433 & 0.336 & 0.538 \\

\multirow[t]{4}{*}{Reflexion~\cite{shinn2023reflexion}} 
& 0.822 & 0.400 & 0.269 & 0.550 \\

\multirow[t]{4}{*}{Self-Debugging~\cite{chen2024teaching}} &
0.367 & 0.256 & 0.232 & 0.539 \\

\multirow[t]{4}{*}{Self-Planning~\cite{jiang2023self}}   & 0.578 & 0.211 & 0.119 & 0.341\\ 

\multirow[t]{4}{*}{Co-STEER (ours)}    & \textbf{0.889} & \textbf{0.611} & \textbf{0.646} & \textbf{0.887} \\
\bottomrule
\end{tabular}}
\\~\\
\caption{Results of method implementation. All the agent workflows are based on GPT-4-turbo.}
\label{exp:codegen-full}
\end{table}

We compare the implementation ability of our proposed Co-STEER with the baseline agent workflows. The experimental results are shown in Table~\ref{exp:codegen-full}. We observe that our proposed Co-STEER significantly outperforms the baseline models on all four evaluation metrics across 27 test cases, demonstrating the overall effectiveness of Co-STEER. We attribute the significant improvement in implementation ability to both the dynamically expanded knowledge and the retrieval mechanism of the Co-STEER agent. Specifically, Table~\ref{tab:method_comparison} showcases that, similar to Co-STEER, both Reflexion and Self-Debugging adopt the feedback from the environment to enhance the implementation ability of models. The difference is that Co-STEER accumulates practical knowledge through environmental feedback and retrieves its experience (knowledge) according to the current situation, while the other two agents merely consider the feedback of the current implementation and neglect their experience. Intuitively, Co-STEER continuously learns domain knowledge from practice and then effectively uses it, which bridges the gap between a junior and senior engineer, thus leading to significant performance gain. We refer the readers to the appendix for more details.

\subsection{Overall Results of \sname{} (RQ2)}
We study the overall performance of Co-STEER and baselines in the \sname{} scenario: Given the limited number of attempts (representing the real-world limited computational resources) and 40 candidate methods consisting of 27 implementable methods and 13 vague and unimplementable methods, models are expected to achieve the optimal overall performance by the collaborative evolving between their scheduling and implementation abilities. The experimental results are shown in Table~\ref{exp:a3d-full}. We obtain the following findings according to the table.

\begin{table}[!h]
  \centering
  \adjustbox{max width=\textwidth}{
    \begin{tabular}{@{}lcccccccc@{}}
      \toprule
      & \multicolumn{4}{c}{\textbf{Top 5}} & \multicolumn{4}{c}{\textbf{Top 10}} \\
      \cmidrule(lr){2-5} \cmidrule(lr){6-9}
      \textbf{Methods} & \textbf{exec.} & \textbf{format} & \textbf{avg. corr.} & \textbf{max. corr.} & \textbf{exec.} & \textbf{format} & \textbf{avg. corr.} & \textbf{max. corr.} \\
      \midrule
      Random Scheduler & 0.522 & 0.400 & 0.211 & 0.444 & 0.567 & 0.289 & 0.417 & 0.655 \\
      Evolving Scheduler & 0.765 & 0.259 & 0.280 & 0.515 & 0.815 & 0.358 & 0.519 & 0.778 \\
      \midrule
      & \multicolumn{4}{c}{\textbf{Top 15}} & \multicolumn{4}{c}{\textbf{Top 20}} \\
      \midrule
      Random Scheduler & 0.856 & 0.544 & 0.594 & 0.778 & 0.911 & 0.589 & 0.532 & 0.778 \\
      Evolving Scheduler & 0.856 & 0.556 & 0.584 & 0.872 & \textbf{0.878} & \textbf{0.567} & \textbf{0.792} & \textbf{0.987} \\
      \bottomrule
    \end{tabular}
  }
  \caption{Results of model performance on \sname{} scenario based on co-evolving strategy.}
  \label{exp:a3d-full}
      \vspace{-0.4cm}
\end{table}

\noindent \textbf{The evolving scheduler of Co-STEER effectively contributes to its overall performance on \sname{}.}
We observe that the performance of Co-STEER equipped with an evolving scheduler significantly outperforms that of Co-STEER equipped with a random scheduler in the same experimental settings. The results demonstrate that 1) our scheduler exerts a significant positive effect on the performance of Co-STEER, and 2) the order of method implementation plays an important role in achieving a better overall performance. Intuitively, accumulating different kinds of practical knowledge forms different engineering insights, thus affecting the performance of agents within a certain implementation order. The agent learns from practical experience and ``knows'' 1) which method is hard to implement and 2) whether the implementation of a method will facilitate the implementation of another method. 

\noindent \textbf{Larger computational resources unlock more potential of evolving-based methods.}
We observe from Table~\ref{exp:a3d-full} that the performance of the Co-STEER agent improves when more computational resources are available, regardless of the scheduler. The experimental results demonstrate the advantage of our proposed evolving strategy and our constructed knowledge. Specifically, computational resources do not affect the performance of baseline self-correction models, since they always focus on the feedback obtained in the current attempt without knowledge accumulation and retrieval. In contrast, given more computational resources, Co-STEER accumulates more practical knowledge and retrieves it when needed, thus forming its evolutionary characteristic. Since the quality of the scheduler affects the practical experience (an easy or tough beginning) of the agent, both the success and failure experiences contribute to the evolution of Co-STEER: Co-STEER evolves through practice (more computational resources) with both random and evolving schedulers.

\subsection{Visualization of the Evolution of Co-STEER}
\begin{figure}[!ht]
    \vspace{-0.3cm}
    \centering
    \includegraphics[scale=0.3]{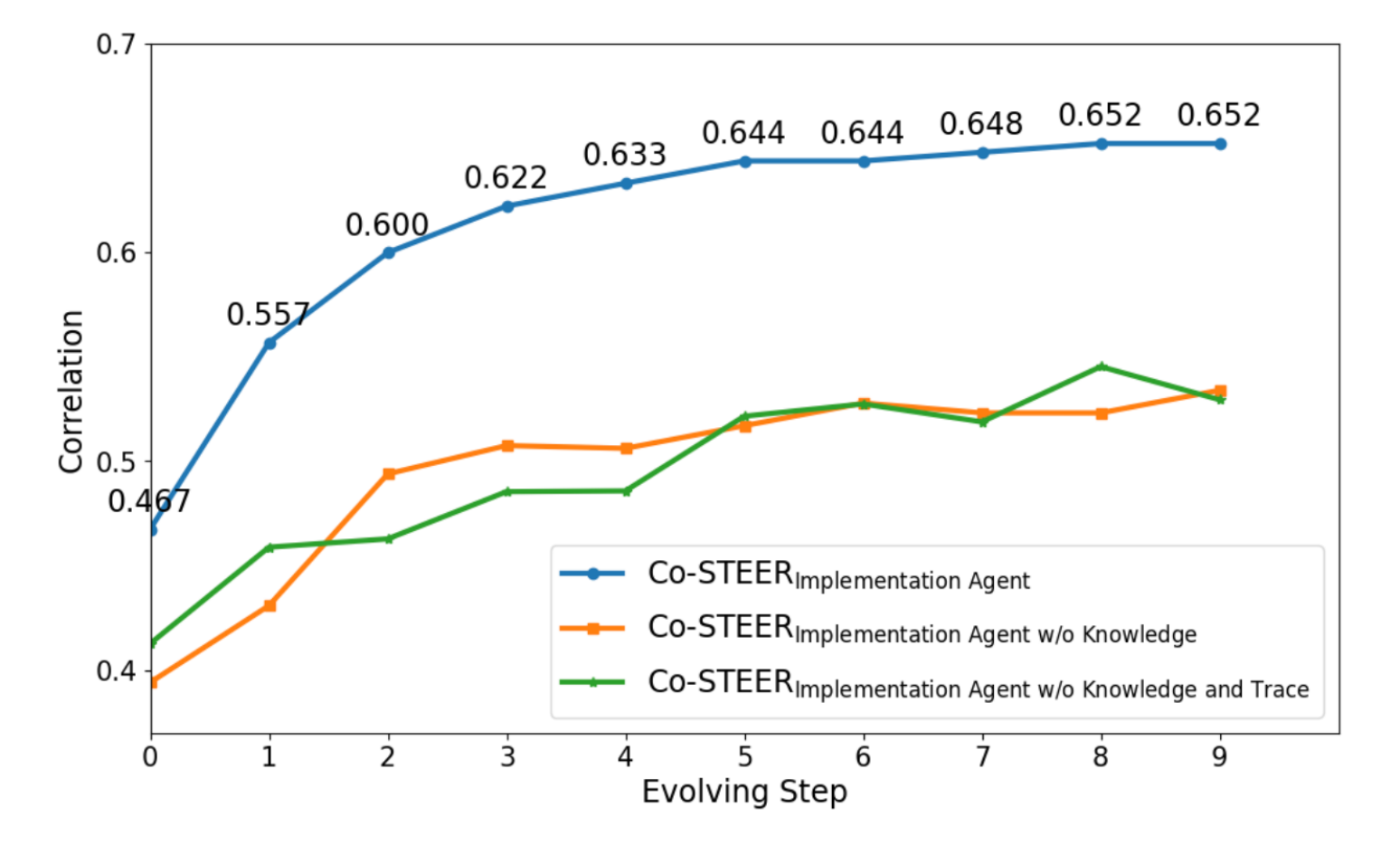}
    \vspace{-0.4cm}
    \caption{Visualization of Co-STEER progress.}
    \label{fig3}
    \vspace{-0.1cm}
\end{figure}

We visualize the evolution of the implementation ability of Co-STEER. As shown in Figure~\ref{fig3}, the implementation ability of Co-STEER has improved compared with its former state in each evolving step. With the proceeding of evolution, Co-STEER gradually approaches its performance boundary. The boundary is decided by the inner intelligence level of the foundation models (e.g., GPT-4) adopted by the current agent. 

We also visualize the self-correction process of baseline models. As mentioned in Table~\ref{tab:method_comparison}, the baseline models with no access to knowledge exhibit marginal performance gain through the self-correction times. Furthermore, without both knowledge and self-correction attempts, a model succeeds in improving itself by merely observing the feedback from the environment.


%
\section{Conclusion}
In this research, we highlight a critical yet overlooked scenario, \sname{}, which holds the potential to significantly advance AI research. Through a detailed examination of this scenario and its associated challenges in method scheduling and implementation, we introduce a groundbreaking autonomous agent, \ppname{}. This agent leverages the power of LLMs and is designed to tackle these challenges. Specifically, it enhances the agent's domain knowledge via a co-evolving strategy and bolsters its scheduling and implementation skills through the accumulation and retrieval of domain-specific experience. Our extensive experiments showcase that the \ppname{} agent, equipped with an evolvable capability of scheduling and implementation, can significantly outperform previous SOTA methods.

\bibliography{main}

\begin{thebibliography}{10}

\bibitem{githubcopilot}
Github copilot: Your ai pair programmer.
\newblock \url{https://github.com/features/copilot}, 2024.

\bibitem{an2024learning}
Shengnan An, Zexiong Ma, Zeqi Lin, Nanning Zheng, Jian-Guang Lou, and Weizhu Chen.
\newblock Learning {{From Mistakes Makes LLM Better Reasoner}}, March 2024.

\bibitem{baek2024researchagent}
Jinheon Baek, Sujay~Kumar Jauhar, Silviu Cucerzan, and Sung~Ju Hwang.
\newblock {{ResearchAgent}}: {{Iterative Research Idea Generation}} over {{Scientific Literature}} with {{Large Language Models}}, April 2024.

\bibitem{bloom2020are}
Nicholas Bloom, Charles~I. Jones, John Van~Reenen, and Michael Webb.
\newblock Are {{Ideas Getting Harder}} to {{Find}}?
\newblock {\em American Economic Review}, 110(4):1104--1144, April 2020.

\bibitem{boiko2023emergent}
Daniil~A Boiko, Robert MacKnight, and Gabe Gomes.
\newblock Emergent autonomous scientific research capabilities of large language models.
\newblock 2023.

\bibitem{brown2020lanugage}
Tom Brown, Benjamin Mann, Nick Ryder, Melanie Subbiah, Jared~D Kaplan, Prafulla Dhariwal, Arvind Neelakantan, Pranav Shyam, Girish Sastry, Amanda Askell, Sandhini Agarwal, Ariel {Herbert-Voss}, Gretchen Krueger, Tom Henighan, Rewon Child, Aditya Ramesh, Daniel Ziegler, Jeffrey Wu, Clemens Winter, Chris Hesse, Mark Chen, Eric Sigler, Mateusz Litwin, Scott Gray, Benjamin Chess, Jack Clark, Christopher Berner, Sam McCandlish, Alec Radford, Ilya Sutskever, and Dario Amodei.
\newblock Language models are few-shot learners.
\newblock In H.~Larochelle, M.~Ranzato, R.~Hadsell, M.F. Balcan, and H.~Lin, editors, {\em Advances in Neural Information Processing Systems}, volume~33, pages 1877--1901. {Curran Associates, Inc.}, 2020.

\bibitem{chen2024rdbench}
Haotian Chen, Xinjie Shen, Zeqi Ye, Xiao Yang, Xu~Yang, Weiqing Liu, and Jiang Bian.
\newblock {{RD2Bench}}: {{Toward}} data-centric automatic {{R}}\&{{D}}.
\newblock In {\em {{ICLR}} 2024 Workshop: {{How}} Far Are We from {{AGI}}}, 2024.

\bibitem{chen2024teaching}
Xinyun Chen, Maxwell Lin, Nathanael Sch{\"a}rli, and Denny Zhou.
\newblock Teaching large language models to self-debug.
\newblock In {\em The Twelfth International Conference on Learning Representations}, 2024.

\bibitem{devlin2018bert}
Jacob Devlin, Ming-Wei Chang, Kenton Lee, and Kristina~N. Toutanova.
\newblock {{BERT}}: {{Pre-training}} of {{Deep Bidirectional Transformers}} for {{Language Understanding}}.
\newblock In {\em Proceedings of the 2019 {{Conference}} of the {{North American Chapter}} of the {{Association}} for {{Computational Linguistics}}: {{Human Language Technologies}}, {{Volume}} 1 ({{Long}} and {{Short Papers}})}, pages 4171--4186, 2018.

\bibitem{durante2024an}
Zane Durante, Bidipta Sarkar, Ran Gong, Rohan Taori, Yusuke Noda, Paul Tang, Ehsan Adeli, Shrinidhi~Kowshika Lakshmikanth, Kevin Schulman, Arnold Milstein, Demetri Terzopoulos, Ade Famoti, Noboru Kuno, Ashley~J. Llorens, Hoi Vo, Katsushi Ikeuchi, Fei-Fei Li, Jianfeng Gao, Naoki Wake, and Qiuyuan Huang.
\newblock An agent foundation model.
\newblock {arXiv}, February 2024.

\bibitem{gao2024clova}
Zhi Gao, Yuntao Du, Xintong Zhang, Xiaojian Ma, Wenjuan Han, Song-Chun Zhu, and Qing Li.
\newblock {{CLOVA}}: {{A Closed-Loop Visual Assistant}} with {{Tool Usage}} and {{Update}}, March 2024.

\bibitem{gou2024critic}
Zhibin Gou, Zhihong Shao, Yeyun Gong, yelong {shen}, Yujiu Yang, Nan Duan, and Weizhu Chen.
\newblock {{CRITIC}}: {{Large}} language models can self-correct with tool-interactive critiquing.
\newblock In {\em The Twelfth International Conference on Learning Representations}, 2024.

\bibitem{hey2009the}
Tony Hey, Stewart Tansley, Kristin Tolle, and Jim Gray.
\newblock {\em The Fourth Paradigm: {{Data-intensive}} Scientific Discovery}.
\newblock {Microsoft Research}, October 2009.

\bibitem{hollmann2024large}
Noah Hollmann, Samuel M{\"u}ller, and Frank Hutter.
\newblock Large language models for automated data science: Introducing caafe for context-aware automated feature engineering.
\newblock {\em Advances in Neural Information Processing Systems}, 36, 2024.

\bibitem{huang2023benchmarking}
Qian Huang, Jian Vora, Percy Liang, and Jure Leskovec.
\newblock Benchmarking large language models as ai research agents.
\newblock {\em arXiv preprint arXiv:2310.03302}, 2023.

\bibitem{jiang2023self}
Xue Jiang, Yihong Dong, Lecheng Wang, Qiwei Shang, and Ge~Li.
\newblock Self-planning code generation with large language model.
\newblock {\em arXiv preprint arXiv:2303.06689}, 2023.

\bibitem{kakushadze2016101}
Zura Kakushadze.
\newblock 101 formulaic alphas.
\newblock {\em Wilmott}, 2016(84):72--81, 2016.

\bibitem{li2024toolaugmented}
Lei Li, Yekun Chai, Shuohuan Wang, Yu~Sun, Hao Tian, Ningyu Zhang, and Hua Wu.
\newblock Tool-augmented reward modeling.
\newblock In {\em The Twelfth International Conference on Learning Representations}, 2024.

\bibitem{liu2024chipnemo}
Mingjie Liu, Teodor-Dumitru Ene, Robert Kirby, Chris Cheng, Nathaniel Pinckney, Rongjian Liang, Jonah Alben, Himyanshu Anand, Sanmitra Banerjee, Ismet Bayraktaroglu, Bonita Bhaskaran, Bryan Catanzaro, Arjun Chaudhuri, Sharon Clay, Bill Dally, Laura Dang, Parikshit Deshpande, Siddhanth Dhodhi, Sameer Halepete, Eric Hill, Jiashang Hu, Sumit Jain, Ankit Jindal, Brucek Khailany, George Kokai, Kishor Kunal, Xiaowei Li, Charley Lind, Hao Liu, Stuart Oberman, Sujeet Omar, Ghasem Pasandi, Sreedhar Pratty, Jonathan Raiman, Ambar Sarkar, Zhengjiang Shao, Hanfei Sun, Pratik~P. Suthar, Varun Tej, Walker Turner, Kaizhe Xu, and Haoxing Ren.
\newblock {{ChipNeMo}}: {{Domain-Adapted LLMs}} for {{Chip Design}}, April 2024.

\bibitem{lu2024self}
Jianqiao Lu, Wanjun Zhong, Wenyong Huang, Yufei Wang, Fei Mi, Baojun Wang, Weichao Wang, Lifeng Shang, and Qun Liu.
\newblock {{SELF}}: {{Language-driven}} self-evolution for large language model, 2024.

\bibitem{ma2023chain}
Xilai Ma, Jing Li, and Min Zhang.
\newblock Chain of {{Thought}} with {{Explicit Evidence Reasoning}} for {{Few-shot Relation Extraction}}.
\newblock In {\em Findings of the {{Association}} for {{Computational Linguistics}}: {{EMNLP}} 2023}, pages 2334--2352, {Singapore}, 2023. {Association for Computational Linguistics}.

\bibitem{mialon2023augmented}
Gr{\'e}goire Mialon, Roberto Dessi, Maria Lomeli, Christoforos Nalmpantis, Ramakanth Pasunuru, Roberta Raileanu, Baptiste Roziere, Timo Schick, Jane {Dwivedi-Yu}, Asli Celikyilmaz, Edouard Grave, Yann LeCun, and Thomas Scialom.
\newblock Augmented language models: A survey.
\newblock {\em Transactions on Machine Learning Research}, 2023.

\bibitem{mock2023ai}
Marissa Mock, Suzanne Edavettal, Christopher Langmead, and Alan Russell.
\newblock {{AI}} can help to speed up drug discovery {\textemdash} but only if we give it the right data.
\newblock {\em Nature}, 621(7979):467--470, September 2023.

\bibitem{openai2023gpt4}
OpenAI.
\newblock {{GPT-4 Technical Report}}, March 2023.

\bibitem{peng2023check}
Baolin Peng, Michel Galley, Pengcheng He, Hao Cheng, Yujia Xie, Yu~Hu, Qiuyuan Huang, Lars Liden, Zhou Yu, Weizhu Chen, and Jianfeng Gao.
\newblock Check {{Your Facts}} and {{Try Again}}: {{Improving Large Language Models}} with {{External Knowledge}} and {{Automated Feedback}}, March 2023.

\bibitem{peng2023semiparametric}
Guangyue Peng, Tao Ge, Si-Qing Chen, Furu Wei, and Houfeng Wang.
\newblock Semiparametric {{Language Models Are Scalable Continual Learners}}, March 2023.

\bibitem{qin2023tool}
Yujia Qin, Shengding Hu, Yankai Lin, Weize Chen, Ning Ding, Ganqu Cui, Zheni Zeng, Yufei Huang, Chaojun Xiao, Chi Han, et~al.
\newblock Tool learning with foundation models.
\newblock {\em arXiv preprint arXiv:2304.08354}, 2023.

\bibitem{qin2024toolllm}
Yujia Qin, Shihao Liang, Yining Ye, Kunlun Zhu, Lan Yan, Yaxi Lu, Yankai Lin, Xin Cong, Xiangru Tang, Bill Qian, Sihan Zhao, Lauren Hong, Runchu Tian, Ruobing Xie, Jie Zhou, Mark Gerstein, dahai {li}, Zhiyuan Liu, and Maosong Sun.
\newblock {{ToolLLM}}: {{Facilitating}} large language models to master 16000+ real-world {{APIs}}.
\newblock In {\em The Twelfth International Conference on Learning Representations}, 2024.

\bibitem{ridnik2024code}
Tal Ridnik, Dedy Kredo, and Itamar Friedman.
\newblock Code {{Generation}} with {{AlphaCodium}}: {{From Prompt Engineering}} to {{Flow Engineering}}, January 2024.

\bibitem{saunders2022selfcritiquing}
William Saunders, Catherine Yeh, Jeff Wu, Steven Bills, Long Ouyang, Jonathan Ward, and Jan Leike.
\newblock Self-critiquing models for assisting human evaluators, June 2022.

\bibitem{schick2023toolformer}
Timo Schick, Jane {Dwivedi-Yu}, Roberto Dessi, Roberta Raileanu, Maria Lomeli, Eric Hambro, Luke Zettlemoyer, Nicola Cancedda, and Thomas Scialom.
\newblock Toolformer: {{Language}} models can teach themselves to use tools.
\newblock In {\em Thirty-Seventh Conference on Neural Information Processing Systems}, 2023.

\bibitem{shinn2023reflexion}
Noah Shinn, Federico Cassano, Ashwin Gopinath, Karthik Narasimhan, and Shunyu Yao.
\newblock Reflexion: Language agents with verbal reinforcement learning.
\newblock In A.~Oh, T.~Neumann, A.~Globerson, K.~Saenko, M.~Hardt, and S.~Levine, editors, {\em Advances in Neural Information Processing Systems}, volume~36, pages 8634--8652. {Curran Associates, Inc.}, 2023.

\bibitem{tao2024report}
Terence Tao and Laura~H. Greene.
\newblock Report on {{Recommendations}} for {{Supercharging Research}}: {{Harnessing Artificial Intelligence}} to {{Meet Global Challenges}}.
\newblock Technical report, April 2024.

\bibitem{tufano2024autodev}
Michele Tufano, Anisha Agarwal, Jinu Jang, Roshanak~Zilouchian Moghaddam, and Neel Sundaresan.
\newblock Autodev: Automated ai-driven development.
\newblock {\em arXiv preprint arXiv:2403.08299}, 2024.

\bibitem{wang2023voyager}
Guanzhi Wang, Yuqi Xie, Yunfan Jiang, Ajay Mandlekar, Chaowei Xiao, Yuke Zhu, Linxi Fan, and Anima Anandkumar.
\newblock Voyager: {{An}} open-ended embodied agent with large language models.
\newblock In {\em Intrinsically-Motivated and Open-Ended Learning Workshop @{{NeurIPS2023}}}, 2023.

\bibitem{wang2023scientific}
Hanchen Wang, Tianfan Fu, Yuanqi Du, Wenhao Gao, Kexin Huang, Ziming Liu, Payal Chandak, Shengchao Liu, Peter Van~Katwyk, Andreea Deac, Anima Anandkumar, Karianne Bergen, Carla~P. Gomes, Shirley Ho, Pushmeet Kohli, Joan Lasenby, Jure Leskovec, Tie-Yan Liu, Arjun Manrai, Debora Marks, Bharath Ramsundar, Le~Song, Jimeng Sun, Jian Tang, Petar Veli{\v c}kovi{\'c}, Max Welling, Linfeng Zhang, Connor~W. Coley, Yoshua Bengio, and Marinka Zitnik.
\newblock Scientific discovery in the age of artificial intelligence.
\newblock {\em Nature}, 620(7972):47--60, August 2023.

\bibitem{wei2022chain}
Jason Wei, Xuezhi Wang, Dale Schuurmans, Maarten Bosma, brian {ichter}, Fei Xia, Ed~Chi, Quoc~V Le, and Denny Zhou.
\newblock Chain-of-thought prompting elicits reasoning in large language models.
\newblock In S.~Koyejo, S.~Mohamed, A.~Agarwal, D.~Belgrave, K.~Cho, and A.~Oh, editors, {\em Advances in Neural Information Processing Systems}, volume~35, pages 24824--24837. {Curran Associates, Inc.}, 2022.

\bibitem{wu2024oscopilot}
Zhiyong Wu, Chengcheng Han, Zichen Ding, Zhenmin Weng, Zhoumianze Liu, Shunyu Yao, Tao Yu, and Lingpeng Kong.
\newblock {{OS-Copilot}}: {{Towards Generalist Computer Agents}} with {{Self-Improvement}}, February 2024.

\bibitem{xu2023lemur}
Yiheng Xu, Hongjin Su, Chen Xing, Boyu Mi, Qian Liu, Weijia Shi, Binyuan Hui, Fan Zhou, Yitao Liu, Tianbao Xie, Zhoujun Cheng, Siheng Zhao, Lingpeng Kong, Bailin Wang, Caiming Xiong, and Tao Yu.
\newblock Lemur: {{Harmonizing Natural Language}} and {{Code}} for {{Language Agents}}, October 2023.

\bibitem{yang2023failures}
Zeyuan Yang, Peng Li, and Yang Liu.
\newblock Failures {{Pave}} the {{Way}}: {{Enhancing Large Language Models}} through {{Tuning-free Rule Accumulation}}.
\newblock In {\em Proceedings of the 2023 {{Conference}} on {{Empirical Methods}} in {{Natural Language Processing}}}, pages 1751--1777, {Singapore}, 2023. {Association for Computational Linguistics}.

\bibitem{yao2023tree}
Shunyu Yao, Dian Yu, Jeffrey Zhao, Izhak Shafran, Tom Griffiths, Yuan Cao, and Karthik Narasimhan.
\newblock Tree of thoughts: {{Deliberate}} problem solving with large language models.
\newblock In A.~Oh, T.~Neumann, A.~Globerson, K.~Saenko, M.~Hardt, and S.~Levine, editors, {\em Advances in Neural Information Processing Systems}, volume~36, pages 11809--11822. {Curran Associates, Inc.}, 2023.

\bibitem{yao2023react}
Shunyu Yao, Jeffrey Zhao, Dian Yu, Nan Du, Izhak Shafran, Karthik~R Narasimhan, and Yuan Cao.
\newblock {{ReAct}}: {{Synergizing}} reasoning and acting in language models.
\newblock In {\em The Eleventh International Conference on Learning Representations}, 2023.

\bibitem{yuan2024craft}
Lifan Yuan, Yangyi Chen, Xingyao Wang, Yi~Fung, Hao Peng, and Heng Ji.
\newblock {{CRAFT}}: {{Customizing LLMs}} by creating and retrieving from specialized toolsets.
\newblock In {\em The Twelfth International Conference on Learning Representations}, 2024.

\bibitem{zeni2024mattergen}
Claudio Zeni, Robert Pinsler, Daniel Z{\"u}gner, Andrew Fowler, Matthew Horton, Xiang Fu, Sasha Shysheya, Jonathan Crabb{\'e}, Lixin Sun, Jake Smith, Bichlien Nguyen, Hannes Schulz, Sarah Lewis, Chin-Wei Huang, Ziheng Lu, Yichi Zhou, Han Yang, Hongxia Hao, Jielan Li, Ryota Tomioka, and Tian Xie.
\newblock {{MatterGen}}: A generative model for inorganic materials design, January 2024.

\bibitem{zha2023data}
Daochen Zha, Zaid~Pervaiz Bhat, Kwei-Herng Lai, Fan Yang, Zhimeng Jiang, Shaochen Zhong, and Xia Hu.
\newblock Data-centric artificial intelligence: A survey.
\newblock {\em arXiv preprint arXiv:2303.10158}, 2023.

\bibitem{zhang2023mlcopilot}
Lei Zhang, Yuge Zhang, Kan Ren, Dongsheng Li, and Yuqing Yang.
\newblock Mlcopilot: Unleashing the power of large language models in solving machine learning tasks.
\newblock {\em arXiv preprint arXiv:2304.14979}, 2023.

\bibitem{zhang2024training}
Shaokun Zhang, Jieyu Zhang, Jiale Liu, Linxin Song, Chi Wang, Ranjay Krishna, and Qingyun Wu.
\newblock Training {{Language Model Agents}} without {{Modifying Language Models}}, February 2024.

\bibitem{zhang2023automatic}
Zhuosheng Zhang, Aston Zhang, Mu~Li, and Alex Smola.
\newblock Automatic chain of thought prompting in large language models.
\newblock In {\em The Eleventh International Conference on Learning Representations}, 2023.

\end{thebibliography}
\bibliographystyle{plain}


\newpage
\appendix


\section{Broader Impacts}
The broader impacts of Co-STEER include accelerating innovation by allowing developers to focus on more creative tasks, economic implications such as potential workforce reductions due to automation, and the need for updated educational programs to equip future workers with relevant skills. Ethical considerations must also be addressed to ensure that such automated systems do not perpetuate biases, especially in sensitive decision-making processes.
\section{Limitation}
The proposed Co-STEER agent showcases significant advancements in automating and optimizing research and development tasks, particularly in financial factor implementation, potentially enhancing R\&D efficiency across various industries. However, its effectiveness heavily relies on the availability of high-quality data and substantial computational resources, which may limit its applicability in resource-constrained environments or other domains without extensive customization. However, such a limitation can be solved in the future as we are having more and more lower price for tokens from LLM.

\section{LLM Prompt Design}
\label{sec:prompt}
\subsection{Scheduling Agent Prompt}
\begin{figure}[h]
    \centering
    \includegraphics[scale=0.4]{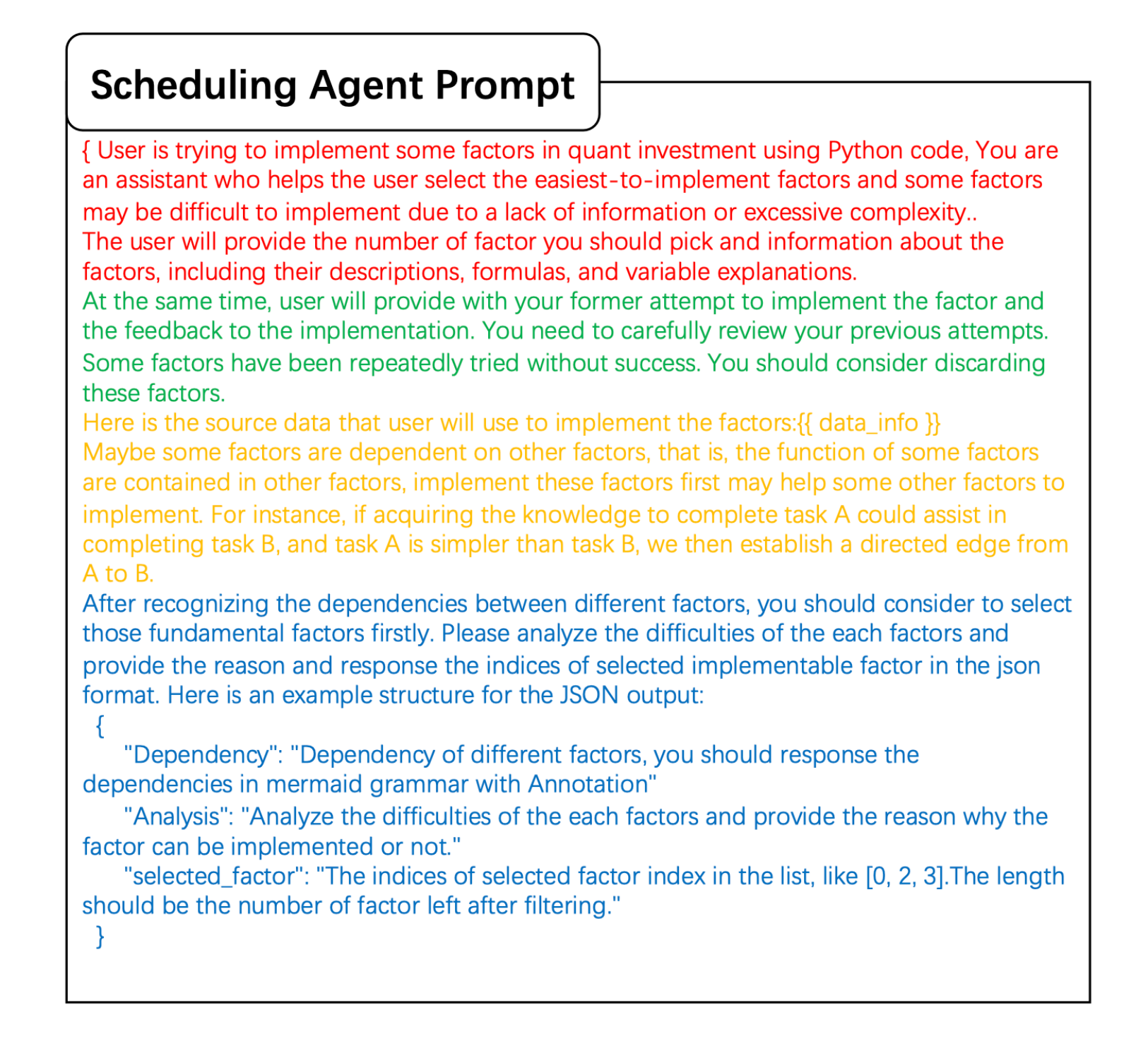}
    \caption{Scheduling agent learn to schedule based on task info \& practical feedbacks.}
    \label{fig:11}
    
\end{figure}
Figure shown in ~\ref{fig:11}.

\subsection{Scheduling Agent Response}
\begin{figure}[h]
    \centering
    \includegraphics[scale=0.4]{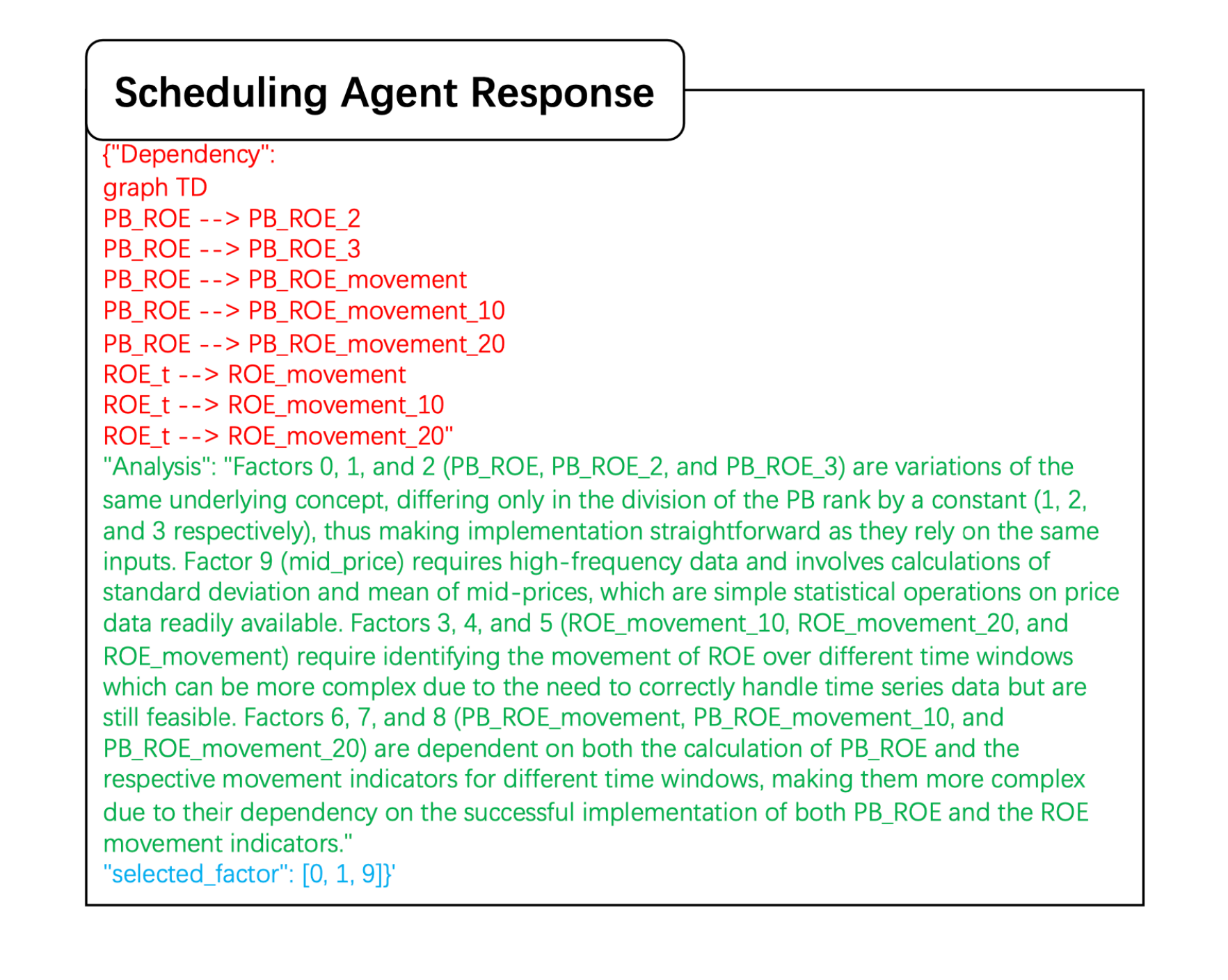}
    \caption{Scheduling agent response: Task complexity and task dependency are considered for multiple factors to prioritize tasks.}
    \label{fig:22}
\end{figure}
Figure shown in ~\ref{fig:22}.

\subsection{Latest attempt with corresponding feedback}
\begin{figure}[h]
    \centering
    \includegraphics[scale=0.6]{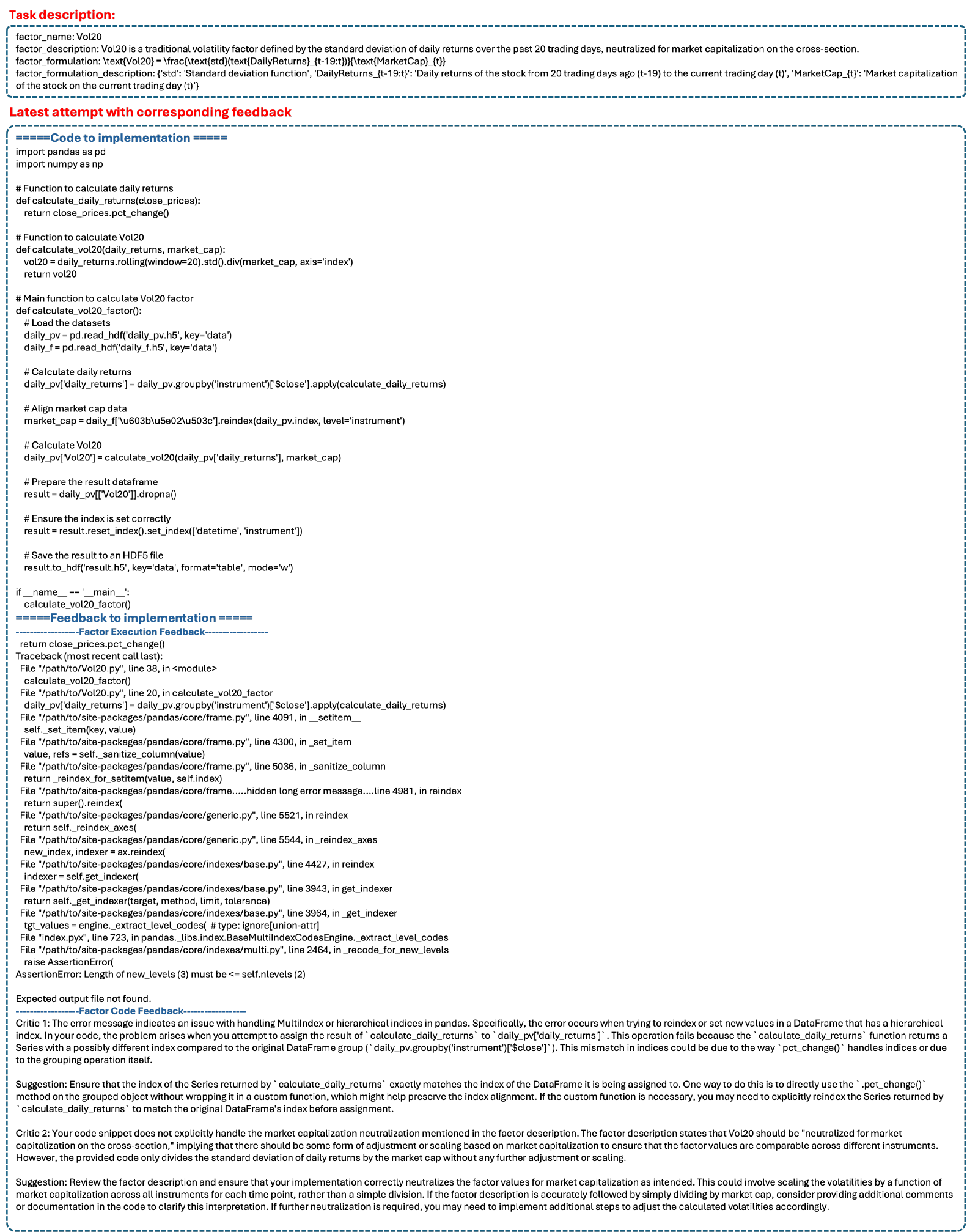}
    \caption{Task description and latest attempt with corresponding feedback.}
    \label{fig:33}
\end{figure}

Figure shown in ~\ref{fig:33}.
\subsection{Retrieved similar correct implementation and error}
\begin{figure}[h]
    \centering
    \includegraphics[scale=0.6]{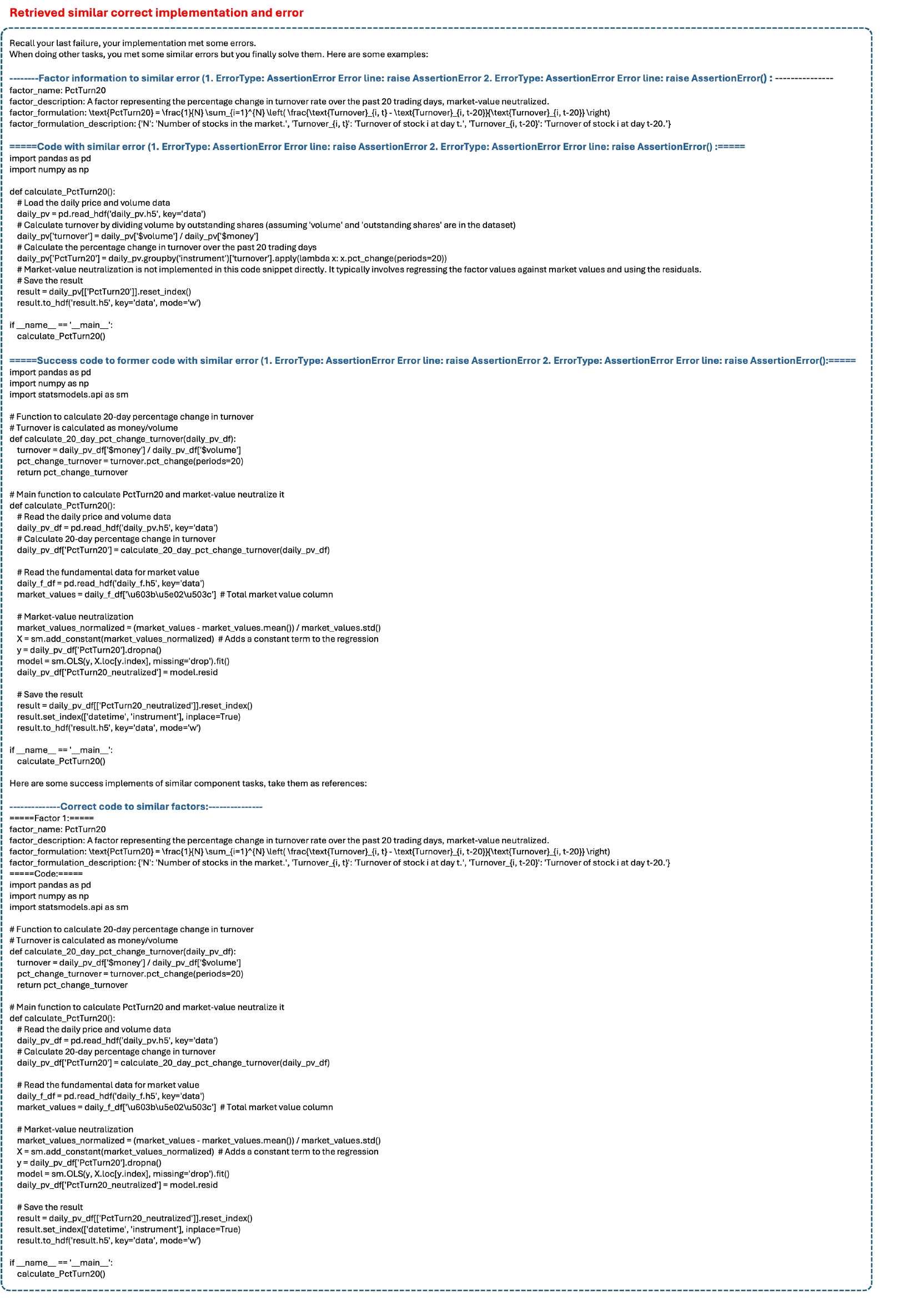}
    \caption{Retrieved similar correct implementation and error}
    \label{fig:44}
\end{figure}
Figure shown in ~\ref{fig:44}.

\section{Experimental Results}
\begin{table}[h]
\centering
\adjustbox{width=1\textwidth}{%
\centering
\begin{tabular}{lllrrrr}
\toprule
 &  &  & avg. exec. & avg. format & avg. corr. & max. corr. \\
\midrule
\multirow[t]{3}{*}{Fundamental} & Easy & PB\_ROE & 0.600 & 0.600 & 0.751 & 1.000 \\
\cline{2-7}
 & Medium & ROE\_movement & 0.200 & 0.100 & 1.000 & 1.000 \\
\cline{2-7}
 & Hard & PB\_ROE\_movement & 0.300 & 0.200 & 0.010 & 0.031 \\
\cline{1-7} \cline{2-7}
\multirow[t]{3}{*}{High Frequency} & Easy & mid\_price & 0.900 & 0.200 & 1.000 & 1.000 \\
\cline{2-7}
 & Medium & liquidity\_imbalance & 0.600 & 0.000 & NaN & NaN \\
\cline{2-7}
 & Hard & micro\_price & 1.000 & 0.300 & 1.000 & 1.000 \\
\cline{1-7} \cline{2-7}
\multirow[t]{3}{*}{Price Volume} & Easy & alpha053 & 1.000 & 0.800 & 0.303 & 1.000 \\
\cline{2-7}
 & Medium & alpha\_pv\_diff & 1.000 & 0.800 & 0.018 & 0.025 \\
\cline{2-7}
 & Hard & alpha\_pv\_diff\_pct & 1.000 & 0.900 & 0.001 & 0.001 \\
\cline{1-7} \cline{2-7}
\multirow[t]{4}{*}{Few-shot} & \multirow[t]{4}{*}{} & Fundamental Avg & 0.367 & 0.300 & 0.587 & 0.677 \\
 &  & High Frequency Avg & 0.833 & 0.167 & 0.667 & 0.667 \\
 &  & Price Volume Avg & 1.000 & 0.833 & 0.107 & 0.342 \\
 &  & mean value  & 0.733 & 0.433 & 0.454 & 0.562 \\
\bottomrule
\end{tabular}}
\space
\caption{Few-shot}
\end{table}

\begin{table}[h]
\adjustbox{width=1\textwidth}{%
\centering
\begin{tabular}{lllrrrr}
\toprule
 &  &  & avg. exec. & avg. format & avg. corr. & max. corr. \\
\midrule
\multirow[t]{3}{*}{Fundamental} & Easy & PB\_ROE & 0.700 & 0.500 & 0.506 & 0.843 \\
\cline{2-7}
 & Medium & ROE\_movement & 0.800 & 0.800 & 0.877 & 1.000 \\
\cline{2-7}
 & Hard & PB\_ROE\_movement & 0.400 & 0.000 & NaN & NaN \\
\cline{1-7} \cline{2-7}
\multirow[t]{3}{*}{High Frequency} & Easy & mid\_price & 0.900 & 0.000 & NaN & NaN \\
\cline{2-7}
 & Medium & liquidity\_imbalance & 0.900 & 0.000 & NaN & NaN \\
\cline{2-7}
 & Hard & micro\_price & 0.900 & 0.000 & NaN & NaN \\
\cline{1-7} \cline{2-7}
\multirow[t]{3}{*}{Price Volume} & Easy & alpha053 & 1.000 & 0.900 & 0.733 & 1.000 \\
\cline{2-7}
 & Medium & alpha\_pv\_diff & 1.000 & 0.900 & 0.133 & 1.000 \\
\cline{2-7}
 & Hard & alpha\_pv\_diff\_pct & 0.900 & 0.800 & 0.778 & 1.000 \\
\cline{1-7} \cline{2-7}
\multirow[t]{4}{*}{CoT} & \multirow[t]{4}{*}{} & Fundamental Avg & 0.633 & 0.433 & 0.461 & 0.614 \\
 &  & High Frequency Avg & 0.900 & 0.000 & 0.000 & 0.000 \\
 &  & Price Volume Avg & 0.967 & 0.867 & 0.548 & 1.000 \\
 &  & mean value (0 for NaN) & 0.833 & 0.433 & 0.336 & 0.538 \\
\bottomrule
\end{tabular}

}
\space
\caption{CoT}
\end{table}

\begin{table}[h]
\adjustbox{width=1\textwidth}{%
\centering
\begin{tabular}{lllrrrr}
\toprule
 &  &  & avg. exec. & avg. format & avg. corr. & max. corr. \\
\midrule
\multirow[t]{3}{*}{Fundamental} & Easy & PB\_ROE & 0.700 & 0.600 & 0.588 & 0.918 \\
\cline{2-7}
 & Medium & ROE\_movement & 1.000 & 0.900 & 0.404 & 1.000 \\
\cline{2-7}
 & Hard & PB\_ROE\_movement & 0.600 & 0.400 & 0.007 & 0.030 \\
\cline{1-7} \cline{2-7}
\multirow[t]{3}{*}{High Frequency} & Easy & mid\_price & 0.700 & 0.000 & NaN & NaN \\
\cline{2-7}
 & Medium & liquidity\_imbalance & 0.700 & 0.000 & NaN & NaN \\
\cline{2-7}
 & Hard & micro\_price & 0.800 & 0.000 & NaN & NaN \\
\cline{1-7} \cline{2-7}
\multirow[t]{3}{*}{Price Volume} & Easy & alpha053 & 1.000 & 0.900 & 0.900 & 1.000 \\
\cline{2-7}
 & Medium & alpha\_pv\_diff & 0.900 & 0.400 & 0.119 & 1.000 \\
\cline{2-7}
 & Hard & alpha\_pv\_diff\_pct & 1.000 & 0.400 & 0.400 & 1.000 \\
\cline{1-7} \cline{2-7}
\multirow[t]{4}{*}{Reflexion} & \multirow[t]{4}{*}{} & Fundamental Avg & 0.767 & 0.633 & 0.333 & 0.649 \\
 &  & High Frequency Avg & 0.733 & 0.000 & 0.000 & 0.000 \\
 &  & Price Volume Avg & 0.967 & 0.567 & 0.473 & 1.000 \\
 &  & mean value (0 for NaN) & 0.822 & 0.400 & 0.269 & 0.550 \\

\bottomrule
\end{tabular}
}
\space
\caption{Reflexion}
\end{table}

\begin{table}[h]
\adjustbox{width=1\textwidth}{%
\centering
\begin{tabular}{lllrrrr}
\toprule
 &  &  & avg. exec. & avg. format & avg. corr. & max. corr. \\
\midrule
\multirow[t]{3}{*}{Fundamental} & Easy & PB\_ROE & 0.500 & 0.500 & 0.033 & 0.953 \\
\cline{2-7}
 & Medium & ROE\_movement & 0.700 & 0.700 & 0.292 & 1.000 \\
\cline{2-7}
 & Hard & PB\_ROE\_movement & 0.900 & 0.700 & 0.318 & 0.898 \\
\cline{1-7} \cline{2-7}
\multirow[t]{3}{*}{High Frequency} & Easy & mid\_price & 0.200 & 0.000 & NaN & NaN \\
\cline{2-7}
 & Medium & liquidity\_imbalance & 0.400 & 0.000 & NaN & NaN \\
\cline{2-7}
 & Hard & micro\_price & 0.200 & 0.000 & NaN & NaN \\
\cline{1-7} \cline{2-7}
\multirow[t]{3}{*}{Price Volume} & Easy & alpha053 & 0.200 & 0.200 & 1.000 & 1.000 \\
\cline{2-7}
 & Medium & alpha\_pv\_diff & 0.000 & 0.000 & NaN & NaN \\
\cline{2-7}
 & Hard & alpha\_pv\_diff\_pct & 0.200 & 0.200 & 0.447 & 1.000 \\
\cline{1-7} \cline{2-7}
\multirow[t]{4}{*}{Self-Debugging} & \multirow[t]{4}{*}{} & Fundamental Avg & 0.700 & 0.633 & 0.214 & 0.950 \\
 &  & High Frequency Avg & 0.267 & 0.000 & 0.000 & 0.000 \\
 &  & Price Volume Avg & 0.133 & 0.133 & 0.482 & 0.667 \\
 &  & mean value (0 for NaN) & 0.367 & 0.256 & 0.232 & 0.539 \\
\bottomrule
\end{tabular}
}
\space
\caption{Self-Debugging}
\end{table}

\begin{table}[h]
\adjustbox{width=1\textwidth}{%
\centering
\begin{tabular}{lllrrrr}
\toprule
 &  &  & avg. exec. & avg. format & avg. corr. & max. corr. \\
\midrule
\multirow[t]{3}{*}{Fundamental} & Easy & PB\_ROE & 0.100 & 0.000 & NaN & NaN \\
\cline{2-7}
 & Medium & ROE\_movement & 0.700 & 0.700 & 0.438 & 1.000 \\
\cline{2-7}
 & Hard & PB\_ROE\_movement & 0.400 & 0.200 & 0.014 & 0.048 \\
\cline{1-7} \cline{2-7}
\multirow[t]{3}{*}{High Frequency} & Easy & mid\_price & 0.400 & 0.000 & NaN & NaN \\
\cline{2-7}
 & Medium & liquidity\_imbalance & 0.400 & 0.000 & NaN & NaN \\
\cline{2-7}
 & Hard & micro\_price & 0.300 & 0.100 & NaN & NaN \\
\cline{1-7} \cline{2-7}
\multirow[t]{3}{*}{Price Volume} & Easy & alpha053 & 1.000 & 0.400 & 0.402 & 1.000 \\
\cline{2-7}
 & Medium & alpha\_pv\_diff & 0.900 & 0.200 & 0.017 & 0.025 \\
\cline{2-7}
 & Hard & alpha\_pv\_diff\_pct & 1.000 & 0.300 & 0.200 & 1.000 \\
\cline{1-7} \cline{2-7}
\multirow[t]{4}{*}{Self-planning} & \multirow[t]{4}{*}{} & Fundamental Avg & 0.400 & 0.300 & 0.151 & 0.349 \\
 &  & High Frequency Avg & 0.367 & 0.033 & 0.000 & 0.000 \\
 &  & Price Volume Avg & 0.967 & 0.300 & 0.206 & 0.675 \\
 &  & mean value (0 for NaN) & 0.578 & 0.211 & 0.119 & 0.341 \\
\bottomrule
\end{tabular}

}
\space
\caption{Self-planning}
\end{table}

\begin{table}[h]
\adjustbox{width=1\textwidth}{%
\centering
\begin{tabular}{lllrrrr}
\toprule
 &  &  & avg. exec. & avg. format & avg. corr. & max. corr. \\
\midrule
\multirow[t]{3}{*}{Fundamental} & Easy & PB\_ROE & 1.000 & 1.000 & 0.571 & 0.983 \\
\cline{2-7}
 & Medium & ROE\_movement & 0.900 & 0.900 & 0.451 & 1.000 \\
\cline{2-7}
 & Hard & PB\_ROE\_movement & 0.900 & 0.900 & 0.317 & 0.897 \\
\cline{1-7} \cline{2-7}
\multirow[t]{3}{*}{High Frequency} & Easy & mid\_price & 1.000 & 0.100 & 1.000 & 1.000 \\
\cline{2-7}
 & Medium & liquidity\_imbalance & 1.000 & 0.300 & 1.000 & 1.000 \\
\cline{2-7}
 & Hard & micro\_price & 0.800 & 0.100 & 0.104 & 0.104 \\
\cline{1-7} \cline{2-7}
\multirow[t]{3}{*}{Price Volume} & Easy & alpha053 & 0.500 & 0.400 & 0.800 & 1.000 \\
\cline{2-7}
 & Medium & alpha\_pv\_diff & 1.000 & 1.000 & 1.000 & 1.000 \\
\cline{2-7}
 & Hard & alpha\_pv\_diff\_pct & 0.900 & 0.800 & 0.573 & 1.000 \\
\cline{1-7} \cline{2-7}
\multirow[t]{4}{*}{Co-STEER (ours)} & \multirow[t]{4}{*}{} & Fundamental Avg & 0.933 & 0.933 & 0.446 & 0.960 \\
 &  & High Frequency Avg & 0.933 & 0.167 & 0.701 & 0.701 \\
 &  & Price Volume Avg & 0.800 & 0.733 & 0.791 & 1.000 \\
 &  & mean Value (0 for NaN) & 0.889 & 0.611 & 0.646 & 0.887 \\
\bottomrule
\end{tabular}
}
\space
\caption{Co-STEER (ours)}
\end{table}

\begin{table}[h]
\centering
\scalebox{0.8}{
\begin{tabular}{llcccc}
\toprule
 Model & Category & Avg. Exec. & Avg. Form. & Avg. Corr. & Max. Corr. \\
\midrule
\multirow[t]{4}{*}{Few-shot~\cite{brown2020lanugage}} & Fundamental & 0.367 & 0.300 & 0.587 & 0.677 \\
  & High Frequency & 0.833 & 0.167 & 0.667 & 0.667 \\
  & Price Volume & 1.000 & 0.833 & 0.107 & 0.342 \\
  & Mean Value & 0.733 & 0.433 & 0.454 & 0.562 \\
\hline
\multirow[t]{4}{*}{CoT~\cite{wei2022chain}} & Fundamental & 0.633 & 0.433 & 0.461 & 0.614 \\
  & High Frequency & 0.900 & 0.000 & 0.000 & 0.000 \\
  & Price Volume & 0.967 & 0.867 & 0.548 & 1.000 \\
  & Mean Value & 0.833 & 0.433 & 0.336 & 0.538 \\
\hline
\multirow[t]{4}{*}{Reflexion~\cite{shinn2023reflexion}} & Fundamental & 0.767 & 0.633 & 0.333 & 0.649 \\
  & High Frequency & 0.733 & 0.000 & 0.000 & 0.000 \\
  & Price Volume & 0.967 & 0.567 & 0.473 & 1.000 \\
 & Mean Value & 0.822 & 0.400 & 0.269 & 0.550 \\
\hline
\multirow[t]{4}{*}{Self-Debugging~\cite{chen2024teaching}} & Fundamental & 0.700 & 0.633 & 0.214 & 0.950 \\
  & High Frequency & 0.267 & 0.000 & 0.000 & 0.000 \\
  & Price Volume & 0.133 & 0.133 & 0.482 & 0.667 \\
  & Mean Value  & 0.367 & 0.256 & 0.232 & 0.539 \\
\hline
\multirow[t]{4}{*}{Self-planning~\cite{jiang2023self}} & Fundamental & 0.400 & 0.300 & 0.151 & 0.349 \\
  & High Frequency & 0.367 & 0.033 & 0.000 & 0.000 \\
  & Price Volume & 0.967 & 0.300 & 0.206 & 0.675 \\
  & Mean Value  & 0.578 & 0.211 & 0.119 & 0.341\\
\hline
\multirow[t]{4}{*}{Co-STEER}&
Fundamental  & 0.933 & 0.933 & 0.446 & 0.960 \\
   & High Frequency & 0.933 & 0.167 & 0.701 & 0.701 \\
   & Price Volume Avg & 0.800 & 0.733 & 0.791 & 1.000 \\
   & Mean Value & \textbf{0.889} & \textbf{0.611} & \textbf{0.646} & \textbf{0.887} \\
\bottomrule
\end{tabular}}
\space
\caption{Results of Model Performance in \sname{} scenario. All results in this table represent the mean values of the corresponding category.}
\end{table}

\begin{table}[h]
\adjustbox{width=1\textwidth}{%
\centering
\begin{tabular}{lllrrrr}
\toprule
 &  &  & avg. exec. & avg. format & avg. corr. & max. corr. \\
\midrule
\multirow[t]{3}{*}{Fundamental} & Easy & PB\_ROE & 0.000 & 0.000 & NaN & NaN \\
\cline{2-7}
 & Medium & ROE\_movement & 1.000 & 1.000 & 0.412 & 1.000 \\
\cline{2-7}
 & Hard & PB\_ROE\_movement & 0.800 & 0.700 & 0.318 & 1.000 \\
\cline{1-7} \cline{2-7}
\multirow[t]{3}{*}{High Frequency} & Easy & mid\_price & 0.300 & 0.000 & NaN & NaN \\
\cline{2-7}
 & Medium & liquidity\_imbalance & 0.400 & 0.100 & NaN & NaN \\
\cline{2-7}
 & Hard & micro\_price & 0.300 & 0.000 & NaN & NaN \\
\cline{1-7} \cline{2-7}
\multirow[t]{3}{*}{Price Volume} & Easy & alpha053 & 0.000 & 0.000 & NaN & NaN \\
\cline{2-7}
 & Medium & alpha\_pv\_diff & 1.000 & 1.000 & 0.610 & 1.000 \\
\cline{2-7}
 & Hard & alpha\_pv\_diff\_pct & 0.900 & 0.800 & 0.556 & 1.000 \\
\cline{1-7} \cline{2-7}
\multirow[t]{4}{*}{Random (5)+Evolving} & \multirow[t]{4}{*}{} & Fundamental Avg & 0.600 & 0.567 & 0.243 & 0.667 \\
 &  & High Frequency Avg & 0.333 & 0.033 & 0.000 & 0.000 \\
 &  & Price Volume Avg & 0.633 & 0.600 & 0.389 & 0.667 \\
 &  & mean Value (0 for NaN) & 0.522 & 0.400 & 0.211 & 0.444 \\

\bottomrule
\end{tabular}

}
\space
\caption{Random (5)+Evolving}
\end{table}

\begin{table}[h]
\adjustbox{width=1\textwidth}{%
\centering
\begin{tabular}{lllrrrr}
\toprule
 &  &  & avg. exec. & avg. format & avg. corr. & max. corr. \\
\midrule
\multirow[t]{3}{*}{Fundamental} & Easy & PB\_ROE & 0.000 & 0.000 & NaN & NaN \\
\cline{2-7}
 & Medium & ROE\_movement & 0.800 & 0.500 & 0.341 & 1.000 \\
\cline{2-7}
 & Hard & PB\_ROE\_movement & 0.900 & 0.800 & 0.281 & 0.897 \\
\cline{1-7} \cline{2-7}
\multirow[t]{3}{*}{High Frequency} & Easy & mid\_price & 0.400 & 0.100 & 1.000 & 1.000 \\
\cline{2-7}
 & Medium & liquidity\_imbalance & 0.600 & 0.100 & 1.000 & 1.000 \\
\cline{2-7}
 & Hard & micro\_price & 0.600 & 0.000 & NaN & NaN \\
\cline{1-7} \cline{2-7}
\multirow[t]{3}{*}{Price Volume} & Easy & alpha053 & 0.200 & 0.000 & NaN & NaN \\
\cline{2-7}
 & Medium & alpha\_pv\_diff & 0.800 & 0.600 & 0.628 & 1.000 \\
\cline{2-7}
 & Hard & alpha\_pv\_diff\_pct & 0.800 & 0.500 & 0.500 & 1.000 \\
\cline{1-7} \cline{2-7}
\multirow[t]{4}{*}{Random (10)+Evolving} & \multirow[t]{4}{*}{} & Fundamental Avg & 0.567 & 0.433 & 0.207 & 0.632 \\
 &  & High Frequency Avg & 0.533 & 0.067 & 0.667 & 0.667 \\
 &  & Price Volume Avg & 0.600 & 0.367 & 0.376 & 0.667 \\
 &  & mean Value (0 for NaN) & 0.567 & 0.289 & 0.417 & 0.655 \\
\bottomrule
\end{tabular}

}
\space
\caption{Random (10)+Evolving}
\end{table}

\begin{table}[h]
\adjustbox{width=1\textwidth}{%
\centering
\begin{tabular}{lllrrrr}
\toprule
 &  &  & avg. exec. & avg. format & avg. corr. & max. corr. \\
\midrule
\multirow[t]{3}{*}{Fundamental} & Easy & PB\_ROE & 1.000 & 1.000 & 0.719 & 1.000 \\
\cline{2-7}
 & Medium & ROE\_movement & 1.000 & 0.800 & 0.666 & 1.000 \\
\cline{2-7}
 & Hard & PB\_ROE\_movement & 0.900 & 0.700 & 0.479 & 1.000 \\
\cline{1-7} \cline{2-7}
\multirow[t]{3}{*}{High Frequency} & Easy & mid\_price & 0.900 & 0.000 & NaN & NaN \\
\cline{2-7}
 & Medium & liquidity\_imbalance & 0.900 & 0.100 & 1.000 & 1.000 \\
\cline{2-7}
 & Hard & micro\_price & 0.700 & 0.000 & NaN & NaN \\
\cline{1-7} \cline{2-7}
\multirow[t]{3}{*}{Price Volume} & Easy & alpha053 & 0.700 & 0.700 & 1.000 & 1.000 \\
\cline{2-7}
 & Medium & alpha\_pv\_diff & 0.800 & 0.800 & 0.756 & 1.000 \\
\cline{2-7}
 & Hard & alpha\_pv\_diff\_pct & 0.800 & 0.800 & 0.721 & 1.000 \\
\cline{1-7} \cline{2-7}
\multirow[t]{4}{*}{Random (15)+Evolving} & \multirow[t]{4}{*}{} & Fundamental Avg & 0.967 & 0.833 & 0.622 & 1.000 \\
 &  & High Frequency Avg & 0.833 & 0.033 & 0.333 & 0.333 \\
 &  & Price Volume Avg & 0.767 & 0.767 & 0.826 & 1.000 \\
 &  & mean Value (0 for NaN) & 0.856 & 0.544 & 0.594 & 0.778 \\
\bottomrule
\end{tabular}

}
\space
\caption{Random (15)+Evolving}
\end{table}

\begin{table}[h]
\adjustbox{width=1\textwidth}{%
\centering
\begin{tabular}{lllrrrr}
\toprule
 &  &  & avg. exec. & avg. format & avg. corr. & max. corr. \\
\midrule
\multirow[t]{3}{*}{Fundamental} & Easy & PB\_ROE & 1.000 & 0.900 & 0.741 & 1.000 \\
\cline{2-7}
 & Medium & ROE\_movement & 1.000 & 0.900 & 0.823 & 1.000 \\
\cline{2-7}
 & Hard & PB\_ROE\_movement & 1.000 & 1.000 & 0.487 & 1.000 \\
\cline{1-7} \cline{2-7}
\multirow[t]{3}{*}{High Frequency} & Easy & mid\_price & 0.900 & 0.100 & NaN & NaN \\
\cline{2-7}
 & Medium & liquidity\_imbalance & 1.000 & 0.100 & 1.000 & 1.000 \\
\cline{2-7}
 & Hard & micro\_price & 0.600 & 0.100 & NaN & NaN \\
\cline{1-7} \cline{2-7}
\multirow[t]{3}{*}{Price Volume} & Easy & alpha053 & 0.800 & 0.600 & 0.669 & 1.000 \\
\cline{2-7}
 & Medium & alpha\_pv\_diff & 0.900 & 0.800 & 0.564 & 1.000 \\
\cline{2-7}
 & Hard & alpha\_pv\_diff\_pct & 1.000 & 0.800 & 0.500 & 1.000 \\
\cline{1-7} \cline{2-7}
\multirow[t]{4}{*}{Random (20)+Evolving} & \multirow[t]{4}{*}{} & Fundamental Avg & 1.000 & 0.933 & 0.684 & 1.000 \\
 &  & High Frequency Avg & 0.833 & 0.100 & 0.333 & 0.333 \\
 &  & Price Volume Avg & 0.900 & 0.733 & 0.578 & 1.000 \\
 &  & Mean Value (0 for NaN) & 0.911 & 0.589 & 0.532 & 0.778 \\
\bottomrule
\end{tabular}

}
\space
\caption{Random (20)+Evolving}
\end{table}

\begin{table}[h]
\adjustbox{width=1\textwidth}{%
\centering
\begin{tabular}{lllrrrr}
\toprule
 &  &  & avg. exec. & avg. format & avg. corr. & max. corr. \\
\midrule
\multirow[t]{3}{*}{Fundamental} & Easy & PB\_ROE & 0.778 & 0.444 & 0.446 & 0.953 \\
\cline{2-7}
 & Medium & ROE\_movement & 0.667 & 0.222 & 0.016 & 0.016 \\
\cline{2-7}
 & Hard & PB\_ROE\_movement & 0.889 & 0.222 & 0.347 & 0.668 \\
\cline{1-7} \cline{2-7}
\multirow[t]{3}{*}{High Frequency} & Easy & mid\_price & 0.556 & 0.000 & NaN & NaN \\
\cline{2-7}
 & Medium & liquidity\_imbalance & 0.667 & 0.000 & NaN & NaN \\
\cline{2-7}
 & Hard & micro\_price & 0.667 & 0.000 & NaN & NaN \\
\cline{1-7} \cline{2-7}
\multirow[t]{3}{*}{Price Volume} & Easy & alpha053 & 0.889 & 0.778 & 0.880 & 1.000 \\
\cline{2-7}
 & Medium & alpha\_pv\_diff & 0.889 & 0.111 & 0.167 & 1.000 \\
\cline{2-7}
 & Hard & alpha\_pv\_diff\_pct & 0.889 & 0.556 & 0.666 & 1.000 \\
\cline{1-7} \cline{2-7}
\multirow[t]{4}{*}{Scheduler (5)+Evolving} & \multirow[t]{4}{*}{} & Fundamental Avg & 0.778 & 0.296 & 0.270 & 0.546 \\
 &  & High Frequency Avg & 0.630 & 0.000 & 0.000 & 0.000 \\
 &  & Price Volume Avg & 0.889 & 0.481 & 0.571 & 1.000 \\
 &  & mean value (0 for NaN) & 0.765 & 0.259 & 0.280 & 0.515 \\
\bottomrule
\end{tabular}
}
\space
\caption{Scheduler (5)+Evolving}
\end{table}

\begin{table}[h]
\adjustbox{width=1\textwidth}{%
\centering
\begin{tabular}{lllrrrr}
\toprule
 &  &  & avg. exec. & avg. format & avg. corr. & max. corr. \\
\midrule
\multirow[t]{3}{*}{Fundamental} & Easy & PB\_ROE & 0.667 & 0.333 & 0.617 & 1.000 \\
\cline{2-7}
 & Medium & ROE\_movement & 0.889 & 0.667 & 0.505 & 1.000 \\
\cline{2-7}
 & Hard & PB\_ROE\_movement & 0.889 & 0.667 & 0.500 & 1.000 \\
\cline{1-7} \cline{2-7}
\multirow[t]{3}{*}{High Frequency} & Easy & mid\_price & 0.889 & 0.000 & NaN & NaN \\
\cline{2-7}
 & Medium & liquidity\_imbalance & 0.778 & 0.000 & NaN & NaN \\
\cline{2-7}
 & Hard & micro\_price & 0.889 & 0.111 & 1.000 & 1.000 \\
\cline{1-7} \cline{2-7}
\multirow[t]{3}{*}{Price Volume} & Easy & alpha053 & 0.556 & 0.444 & 1.000 & 1.000 \\
\cline{2-7}
 & Medium & alpha\_pv\_diff & 0.889 & 0.667 & 0.718 & 1.000 \\
\cline{2-7}
 & Hard & alpha\_pv\_diff\_pct & 0.889 & 0.333 & 0.333 & 1.000 \\
\cline{1-7} \cline{2-7}
\multirow[t]{4}{*}{Scheduler (10) + Evolving} & \multirow[t]{4}{*}{} & Fundamental Avg & 0.815 & 0.556 & 0.541 & 1.000 \\
 &  & High Frequency Avg & 0.852 & 0.037 & 0.333 & 0.333 \\
 &  & Price Volume Avg & 0.778 & 0.481 & 0.684 & 1.000 \\
 &  & mean value (0 for NaN) & 0.815 & 0.358 & 0.519 & 0.778 \\
\bottomrule
\end{tabular}
}
\space
\caption{Scheduler (10) + Evolving}
\end{table}

\begin{table}[h]
\adjustbox{width=1\textwidth}{%
\centering
\begin{tabular}{lllrrrr}
\toprule
 &  &  & avg. exec. & avg. format & avg. corr. & max. corr. \\
\midrule
\multirow[t]{3}{*}{Fundamental} & Easy & PB\_ROE & 0.700 & 0.700 & 0.714 & 0.953 \\
\cline{2-7}
 & Medium & ROE\_movement & 0.900 & 0.900 & 0.672 & 1.000 \\
\cline{2-7}
 & Hard & PB\_ROE\_movement & 0.900 & 0.800 & 0.120 & 0.897 \\
\cline{1-7} \cline{2-7}
\multirow[t]{3}{*}{High Frequency} & Easy & mid\_price & 0.800 & 0.000 & NaN & NaN \\
\cline{2-7}
 & Medium & liquidity\_imbalance & 1.000 & 0.100 & 1.000 & 1.000 \\
\cline{2-7}
 & Hard & micro\_price & 0.800 & 0.100 & 1.000 & 1.000 \\
\cline{1-7} \cline{2-7}
\multirow[t]{3}{*}{Price Volume} & Easy & alpha053 & 0.900 & 0.800 & 0.948 & 1.000 \\
\cline{2-7}
 & Medium & alpha\_pv\_diff & 0.700 & 0.700 & 0.304 & 1.000 \\
\cline{2-7}
 & Hard & alpha\_pv\_diff\_pct & 1.000 & 0.900 & 0.500 & 1.000 \\
\cline{1-7} \cline{2-7}
\multirow[t]{4}{*}{Scheduler (15) + Evolving} & \multirow[t]{4}{*}{} & Fundamental Avg & 0.833 & 0.800 & 0.502 & 0.950 \\
 &  & High Frequency Avg & 0.867 & 0.067 & 0.667 & 0.667 \\
 &  & Price Volume Avg & 0.867 & 0.800 & 0.584 & 1.000 \\
 &  & mean value (0 for NaN) & 0.856 & 0.556 & 0.584 & 0.872 \\
\bottomrule
\end{tabular}}
\space
\caption{Scheduler (15) + Evolving}
\end{table}

\begin{table}[h]
\adjustbox{width=1\textwidth}{%
\centering
\begin{tabular}{lllrrrr}
\toprule
 &  &  & avg. exec. & avg. format & avg. corr. & max. corr. \\
\midrule
\multirow[t]{3}{*}{Fundamental} & Easy & PB\_ROE & 0.700 & 0.700 & 0.917 & 0.983 \\
\cline{2-7}
 & Medium & ROE\_movement & 1.000 & 1.000 & 0.701 & 1.000 \\
\cline{2-7}
 & Hard & PB\_ROE\_movement & 0.900 & 0.800 & 0.111 & 0.897 \\
\cline{1-7} \cline{2-7}
\multirow[t]{3}{*}{High Frequency} & Easy & mid\_price & 0.900 & 0.100 & 1.000 & 1.000 \\
\cline{2-7}
 & Medium & liquidity\_imbalance & 1.000 & 0.100 & 1.000 & 1.000 \\
\cline{2-7}
 & Hard & micro\_price & 0.700 & 0.200 & 1.000 & 1.000 \\
\cline{1-7} \cline{2-7}
\multirow[t]{3}{*}{Price Volume} & Easy & alpha053 & 0.700 & 0.700 & 1.000 & 1.000 \\
\cline{2-7}
 & Medium & alpha\_pv\_diff & 1.000 & 0.700 & 0.603 & 1.000 \\
\cline{2-7}
 & Hard & alpha\_pv\_diff\_pct & 1.000 & 0.800 & 0.800 & 1.000 \\
\cline{1-7} \cline{2-7}
\multirow[t]{4}{*}{Scheduler (20) + Evolving} & \multirow[t]{4}{*}{} & Fundamental Avg & 0.867 & 0.833 & 0.576 & 0.960 \\
 &  & High Frequency Avg & 0.867 & 0.133 & 1.000 & 1.000 \\
 &  & Price Volume Avg & 0.900 & 0.733 & 0.801 & 1.000 \\
 &  & mean value (0 for NaN) & 0.878 & 0.567 & 0.792 & 0.987 \\
\bottomrule
\end{tabular}}
\space
\caption{Scheduler (20) + Evolving}
\end{table}

\begin{table}[h]
\centering
\adjustbox{width=0.8\textwidth}{%
\begin{tabular}{llcccc}
\toprule
  Model & Category & Avg. Exec. & Avg. Form. & Avg. Corr. & Max. Corr. \\
\midrule
 \multirow[t]{4}{*}{Random (5)} & Fundamental & 0.600 & 0.567 & 0.243 & 0.667 \\
   & High Frequency & 0.333 & 0.033 & 0.000 & 0.000 \\
  & Price Volume & 0.633 & 0.600 & 0.389 & 0.667 \\
  & Mean Value & 0.522 & 0.400 & 0.211 & 0.444 \\
\hline
\multirow[t]{4}{*}{Random (10)} & Fundamental  & 0.567 & 0.433 & 0.207 & 0.632 \\
  & High Frequency  & 0.533 & 0.067 & 0.667 & 0.667 \\
  & Price Volume  & 0.600 & 0.367 & 0.376 & 0.667 \\
  & Mean Value  & 0.567 & 0.289 & 0.417 & 0.655 \\
\hline
\multirow[t]{4}{*}{Random (15)} & Fundamental  & 0.967 & 0.833 & 0.622 & 1.000 \\
  & High Frequency  & 0.833 & 0.033 & 0.333 & 0.333 \\
  & Price Volume  & 0.767 & 0.767 & 0.826 & 1.000 \\
  & Mean Value  & 0.856 & 0.544 & 0.594 & 0.778 \\
\hline
\multirow[t]{4}{*}{Random (20)} & Fundamental  & 1.000 & 0.933 & 0.684 & 1.000 \\
  & High Frequency  & 0.833 & 0.100 & 0.333 & 0.333 \\
  & Price Volume  & 0.900 & 0.733 & 0.578 & 1.000 \\
  & Mean Value  & 0.911 & 0.589 & 0.532 & 0.778 \\
\hline
\multirow[t]{4}{*}{Scheduler (5)} & Fundamental  & 0.778 & 0.296 & 0.270 & 0.546 \\
  & High Frequency  & 0.630 & 0.000 & 0.000 & 0.000 \\
  & Price Volume  & 0.889 & 0.481 & 0.571 & 1.000 \\
  & Mean Value  & 0.765 & 0.259 & 0.280 & 0.515 \\
\hline
\multirow[t]{4}{*}{Scheduler (10)} & Fundamental  & 0.815 & 0.556 & 0.541 & 1.000 \\
  & High Frequency  & 0.852 & 0.037 & 0.333 & 0.333 \\
  & Price Volume  & 0.778 & 0.481 & 0.684 & 1.000 \\
  & Mean Value  & 0.815 & 0.358 & 0.519 & 0.778 \\
\hline
\multirow[t]{4}{*}{Scheduler (15)} & Fundamental  & 0.833 & 0.800 & 0.502 & 0.950 \\
  & High Frequency  & 0.867 & 0.067 & 0.667 & 0.667 \\
  & Price Volume  & 0.867 & 0.800 & 0.584 & 1.000 \\
  & Mean Value  & 0.856 & 0.556 & 0.584 & 0.872 \\
\hline
\multirow[t]{4}{*}{Scheduler (20)} & Fundamental  & 0.867 & 0.833 & 0.576 & 0.960 \\
  & High Frequency  & 0.867 & 0.133 & 1.000 & 1.000 \\
  & Price Volume  & 0.900 & 0.733 & 0.801 & 1.000 \\
  & Mean Value  & \textbf{0.878} & \textbf{0.567} & \textbf{0.792} & \textbf{0.987} \\
\hline
\end{tabular}}
\space
\caption{Results of model performance on \sname{} scenario by co-evolving. The number in parentheses denotes the number of methods to be implemented. The methods are decided by the scheduler. }
\end{table}

\end{document}